\documentclass{article}

\usepackage{arxiv}

\usepackage[utf8]{inputenc} 
\usepackage[T1]{fontenc}    
\usepackage{hyperref}       
\usepackage{url}            
\usepackage{booktabs}       
\usepackage{amsfonts}       
\usepackage{nicefrac}       
\usepackage{microtype}      
\usepackage{lipsum}		
\usepackage{graphicx}
\usepackage{natbib}
\usepackage{doi}
\usepackage{xcolor}
\usepackage{amsmath}
\usepackage{tabularx}
\usepackage{makecell}

\definecolor{bestgreen}{RGB}{0,100,0}
\definecolor{secondgreen}{RGB}{46,150,79}
\definecolor{thirdgreen}{RGB}{140,198,110}

\newcommand{\best}[1]{\textbf{\textcolor{bestgreen}{#1}}}
\newcommand{\second}[1]{\textcolor{secondgreen}{#1}}
\newcommand{\third}[1]{\textcolor{thirdgreen}{#1}}

\title{Investigating Inductive Biases for Machine Learning Emulation of Sudden Stratospheric Warmings in Idealised Isca Simulations}

\author{
\href{https://orcid.org/0009-0003-5439-0139}{\includegraphics[scale=0.06]{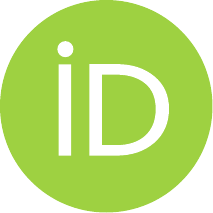}\hspace{1mm}Oskar Bohn Lassen}\\
Department of Technology,\\ Management, and Economics\\
Technical University of Denmark\\
\texttt{obola@dtu.dk}
\And
\href{https://orcid.org/0000-0001-5080-1234}{\includegraphics[scale=0.06]{orcid.pdf}\hspace{1mm}Simon Driscoll}\\
Department of Applied Mathematics\\ and Theoretical Physics\\
University of Cambridge\\
\texttt{sd2136@cam.ac.uk}
\And
\href{https://orcid.org/0000-0002-4775-3259}{\includegraphics[scale=0.06]{orcid.pdf}\hspace{1mm}Stephen I. Thomson}\\
Department of Mathematics\\ and Statistics\\
University of Exeter\\
\texttt{s.i.thomson@exeter.ac.uk}
\And
\href{https://orcid.org/0000-0002-1601-5683}{\includegraphics[scale=0.06]{orcid.pdf}\hspace{1mm}Sebastian Schemm}\\
Department of Applied Mathematics\\ and Theoretical Physics\\
University of Cambridge\\
\texttt{ss3299@cam.ac.uk}
\And
\href{https://orcid.org/0000-0001-5457-9909}{\includegraphics[scale=0.06]{orcid.pdf}\hspace{1mm}Francisco C. Pereira}\\
Department of Technology, \\Management, and Economics\\
Technical University of Denmark\\
\texttt{camara@dtu.dk}
}

\date{}

\renewcommand{\shorttitle}{}

\hypersetup{
pdftitle={Investigating Inductive Biases for Machine Learning Emulation of Sudden Stratospheric Warmings in Idealised Isca Simulations},
pdfsubject={Machine learning, climate emulation, atmospheric dynamics},
pdfauthor={Oskar Bohn Lassen, Simon Driscoll, Stephen I. Thomson, Sebastian Schemm, Francisco C. Pereira},
pdfkeywords={Machine learning, climate emulation, atmospheric dynamics, stratospheric variability, sudden stratospheric warming}
}

\begin{document}
\maketitle

\begin{abstract}
Machine-learning emulators are increasingly used for weather prediction and have the potential to extend skill on subseasonal-to-seasonal timescales by learning dynamically important sources of predictability.
A key challenge is whether the models can exploit predictability anchors, such as stratospheric variability, that influence tropospheric circulation beyond short lead times.
We test how architectural inductive bias affects emulation of sudden stratospheric warming (SSW) dynamics using paired idealised Isca simulations that differ only in an imposed wave-2 heating perturbation. 
Across convolutional, transformer, and graph-based architectures trained for one-step prediction, model differences are modest when the stratosphere is dynamically quiet but widen substantially when SSW-like variability is active. 
Our results identify explicit three-dimensional vertical coupling as a key inductive bias for machine-learning emulation of stratospheric dynamics.
However, Eliassen--Palm flux diagnostics show that low forecast error does not guarantee physically faithful wave--mean-flow interaction, with coherent errors remaining in stratospheric wave-driving structure. 
\end{abstract}


\section{Introduction}
Physics-based general circulation models (GCMs) have been used for decades in weather prediction and climate simulation.
These models evolve atmospheric states through numerical integration of governing equations for large-scale dynamics, together with parameterised representations of unresolved processes~\cite{Stensrud2011, revolution_numerical_weather, futuresofmodeling}.
Because model fidelity and computational cost depend on resolution and on the treatment of subgrid processes, atmospheric scientists use model hierarchies, ranging from comprehensive operational Earth System Models (ESMs) to idealised dynamical-core configurations~\cite{held1994dynamicalcores, modelhierarchies, model_hierarchies_2}.

Despite sustained progress in numerical modelling, the cost of high-resolution simulation and the limitations of physics-based parameterisations have motivated machine-learning (ML) approaches aimed at accelerating and improving atmospheric models~\cite{dueben2018challenges,kashinath2021physicsinformed}.
One prominent direction is hybrid physics-ML modelling, in which ML components emulate or replace expensive parameterisations~\cite{rasp2018deeplearning,brenowitz2018prognostic,yuval2020stable,wang2022stable,bertoli_2025}.
A closely related line of work develops end-to-end hybrid neural GCMs that combine a dynamical solver with learned components in a differentiable framework~\cite{kochkov2024neuralgcm}.
In parallel, purely data-driven models trained directly on reanalysis have demonstrated competitive medium-range forecast skill, including graph-based architectures such as GraphCast, Earth-specific transformers as Pangu-Weather, spectral/Fourier-inspired approaches as FourCastNet, and operational systems as AIFS~\cite{lam2023graphcast,bi2023pangu,pathak2022fourcastnet, Moldovan_2026}.
Foundation models such as Aurora have extended this paradigm by pretraining on Earth-system datasets and fine-tuning to downstream forecasting tasks~\cite{bodnar2025foundation}.
A related line of work is simulator emulation, where a ML model is trained on outputs of a given numerical simulator to learn its state-transition map~\cite{scher2019forecasting, gunawardena_2021_ml_surrogate_atmospheric_transport}. 

However, accurately modelling the stratosphere and its influence on the troposphere remains challenging for ML systems~\cite{lam2023graphcast,bodnar2025foundation, ozdemir_2026} with recent evaluations finding that ML weather models struggle to capture stratosphere-troposphere coupling during sudden stratospheric warmings (SSWs)~\cite{wu2026graphcast_strattrop}. 
SSWs are large disruptions of the polar vortex caused by planetary-wave forcing and wave-mean-flow interaction, and represent an example of complex stratospheric dynamics~\cite{matsuno1971ssw,baldwin2021sswreview}. 
Stratospheric anomalies from SSWs can descend and project onto tropospheric circulation regimes, affecting surface weather on subseasonal timescales \cite{baldwin2001harbingers,kidston2015stratospheric}. 
Hence, stratospheric variability is an important source of subseasonal predictability, but exploiting this source in ML models depends in part on accurately representing vertically propagating waves and wave-mean-flow interaction.
This motivates our central hypothesis: ML architectures that more naturally encode vertically coupled, multiscale dynamics should show relatively larger gains when stratospheric wave dynamics are active. 
Because observational records confound many factors, we use an idealised Isca GCM configuration in which SSW-like variability can be either suppressed or enhanced through modifications to the imposed forcing of the stratosphere by tropospheric wave sources~\cite{jucker2014ssw,mudhar2024modeldependency}.

We generate paired datasets with and without frequent SSW-like disturbances, holding all other model settings fixed.
We then train a suite of ML emulator architectures separately for each regime, spanning 2D and 3D CNNs, transformers, and graph-based models. 
We compare aggregate errors, vertical and latitudinal error structure, and EP-flux diagnostics to test whether vertical-coupling and wave-propagation biases are most beneficial under active SSW-like dynamics.
Using single-timestep diagnostics isolates architectural inductive bias from the rollout-training strategies used in current ML weather models to achieve long-lead skill~\cite{lam2023graphcast}.
Ultimately, this study aims to clarify which design choices are most important for faithfully capturing stratospheric dynamics in ML weather and climate emulation.

\section{Methodology and experimental setup}

\subsection{Isca dynamics \& Eliassen-Palm fluxes}

\begin{figure}
    \centering
    \includegraphics[width=0.9\linewidth]{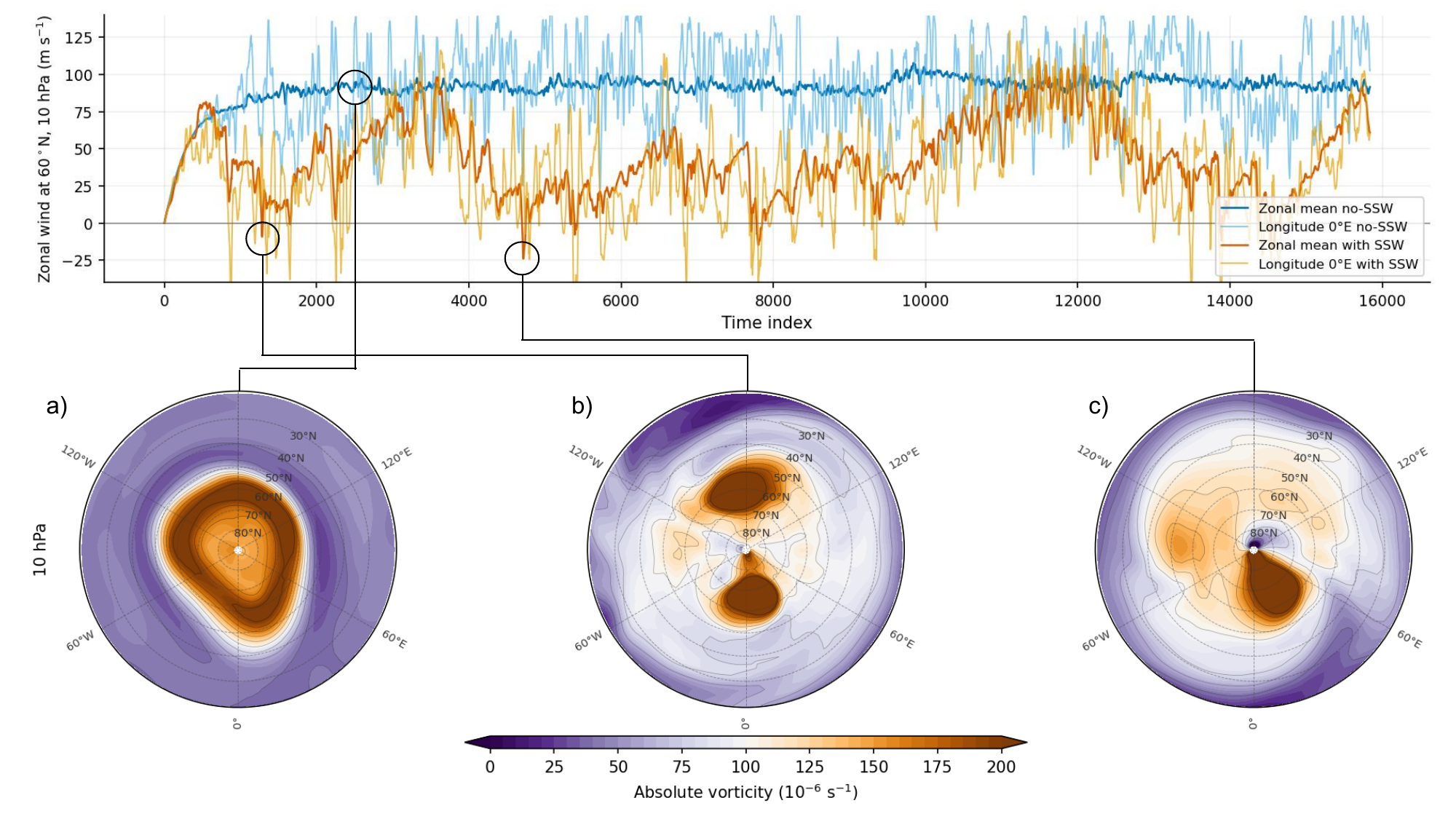}
    \caption{The upper panel shows zonal wind at 60$^\circ$N and 10 hPa for the no-SSW and SSW-enabled configurations.
    Highlighted time steps are shown in the lower panel as polar maps of absolute vorticity: (a) a compact vortex from the no-SSW simulation, (b) a split SSW event from the SSW-enabled simulation, and (c) a displacement SSW event from the SSW-enabled simulation.}
    \label{fig:ssw_examples}
\end{figure}

We use the Isca atmospheric modelling framework to generate controlled idealised simulations of stratospheric variability~\cite{vallis2018isca}. 
Isca is well suited for this purpose because it allows individual physical mechanisms to be isolated in simplified dynamical-core configurations. 
Here, the atmospheric state $\mathbf{x}_t$ contains temperature $T$, zonal wind $u$, and meridional wind $v$ on a latitude--longitude--pressure grid. 
The simulator defines a discrete time-evolution operator,
\begin{equation}
    \mathbf{x}_{t+\Delta t}=\mathcal{M}(\mathbf{x}_t),
\end{equation}
which the machine-learning emulator will be trained to approximate.

The simulations use a dry spectral dynamical-core configuration based on \citet{mudhar2024modeldependency}, with T42 horizontal resolution, 60 vertical pressure levels (38 above 100 hPa), and a 240\,s atmospheric time step.
The troposphere follows a Held--Suarez-style setup, with temperature relaxed toward a prescribed equilibrium profile and low-level winds damped by Rayleigh friction~\cite{held1994dynamicalcores}. 
The stratosphere follows a modified Polvani--Kushner equilibrium-temperature formulation~\cite{polvani2002stratosphere}. 
This means the Isca configuration has no explicit radiation scheme, moisture, or sea-surface temperature as these processess are represented implicitly by the relaxation profile.
The simulations are run in perpetual-winter configuration, such that one hemisphere remains in winter throughout the integration meaning that our 6 years of run-time is effectively 20 winters. 
To generate SSW-like variability, we activate a prescribed zonal wave-2 heating perturbation constant in time, centered near 45$^\circ$N and 400 hPa, following \citet{mudhar2024modeldependency} and \citet{Lindgren_2018}. 
This forcing generates stationary planetary-scale Rossby waves that propagate upward into the stratosphere and disturb the winter polar vortex. 
The paired no-SSW and SSW-enabled datasets are identical except for this imposed wave-heating perturbation.

This paired design is intended as a controlled test of emulator behaviour under two dynamical regimes, rather than as a climatological sampling study of real-world SSW diversity. 
The three-hourly output provides more than 15${,}$000 supervised one-step state transitions, while comparing the two experiments isolates a specific mechanism: dry planetary-wave forcing, upward wave propagation, and polar-vortex disruption. 
By excluding additional sources of variability such as the quasi-biennial oscillation (QBO), volcanic forcing, solar variability, topography, and moist processes, the setup makes it possible to attribute differences in emulator performance directly to the presence of SSW-like wave--mean-flow dynamics.
Using the standard zonal-mean zonal-wind reversal criterion at 60$^\circ$N and 10 hPa, the no-SSW dataset contains no detected SSW events, while the SSW-enabled dataset has an SSW frequency of 0.32 events per 100 model days~\cite{andrew_2007}.

Figure~\ref{fig:ssw_examples} places selected vortex snapshots within the full simulated time series. 
The no-SSW configuration maintains a strong, positive zonal-mean zonal wind at 60$^\circ$N and 10 hPa, indicating a persistent polar vortex, while the SSW-enabled configuration exhibits vortex weakening and reversals. 
The selected maps show representative 10 hPa vortex morphologies: a compact vortex in the no-SSW simulation, and split and displacement events in the SSW-enabled simulation. 
For both simulations, the first 1000 three-hourly output steps are discarded as spin-up, after which the trajectories are converted into supervised one-step prediction pairs $(\mathbf{x}_t,\mathbf{x}_{t+\Delta t})$ with $\Delta t=3$ hours. 
The data are split chronologically into training ($N=2000$), validation ($N=250$), and test ($N=250$) periods, and normalization statistics are computed from the training set only, separately for each variable and pressure level. 
The training set size was chosen as a feasible compromise for large hyperparameter sweeps across architectures, with the data-scaling experiment in the Supporting Information showing that performance continued to improve but with diminishing returns at larger sample sizes.

To evaluate whether the emulators capture physically meaningful wave dynamics, we use Eliassen--Palm flux diagnostics~\cite{andrews_1987}. 
These diagnostics summarize wave--mean-flow interaction through eddy covariance terms $\overline{u'v'}$ and $\overline{v'T'}$, where primes denote deviations from the zonal mean and overbars denote zonal averaging.
In the quasi-geostrophic approximation, the EP-flux vector $\mathbf{F}=(F_{\phi},F_p)$ is proportional to these eddy fluxes, with
\begin{equation}
    F_{\phi}\propto -\overline{u'v'},
    \qquad
    F_p\propto \overline{v'T'}.
\end{equation}
Its divergence, $\nabla\cdot\mathbf{F}$, indicates where wave activity interacts with and forces the zonal-mean flow. 
Because SSW development is associated with enhanced upward and poleward wave propagation into the winter stratosphere, EP-flux diagnostics provide a mechanism-aware complement to pointwise forecast error. 
They test whether an emulator captures not only the local values of $T$, $u$, and $v$, but also the covariance structure responsible for realistic wave--mean-flow interaction. 
A more elaborate explanation of the EP-flux diagnostics in spherical coordinates and Isca configuration is found in the Supporting Information.

\subsection{Machine-Learning Emulation}
\label{sec:ml}

\begin{figure}[t]
    \centering
    \includegraphics[width=0.98\linewidth]{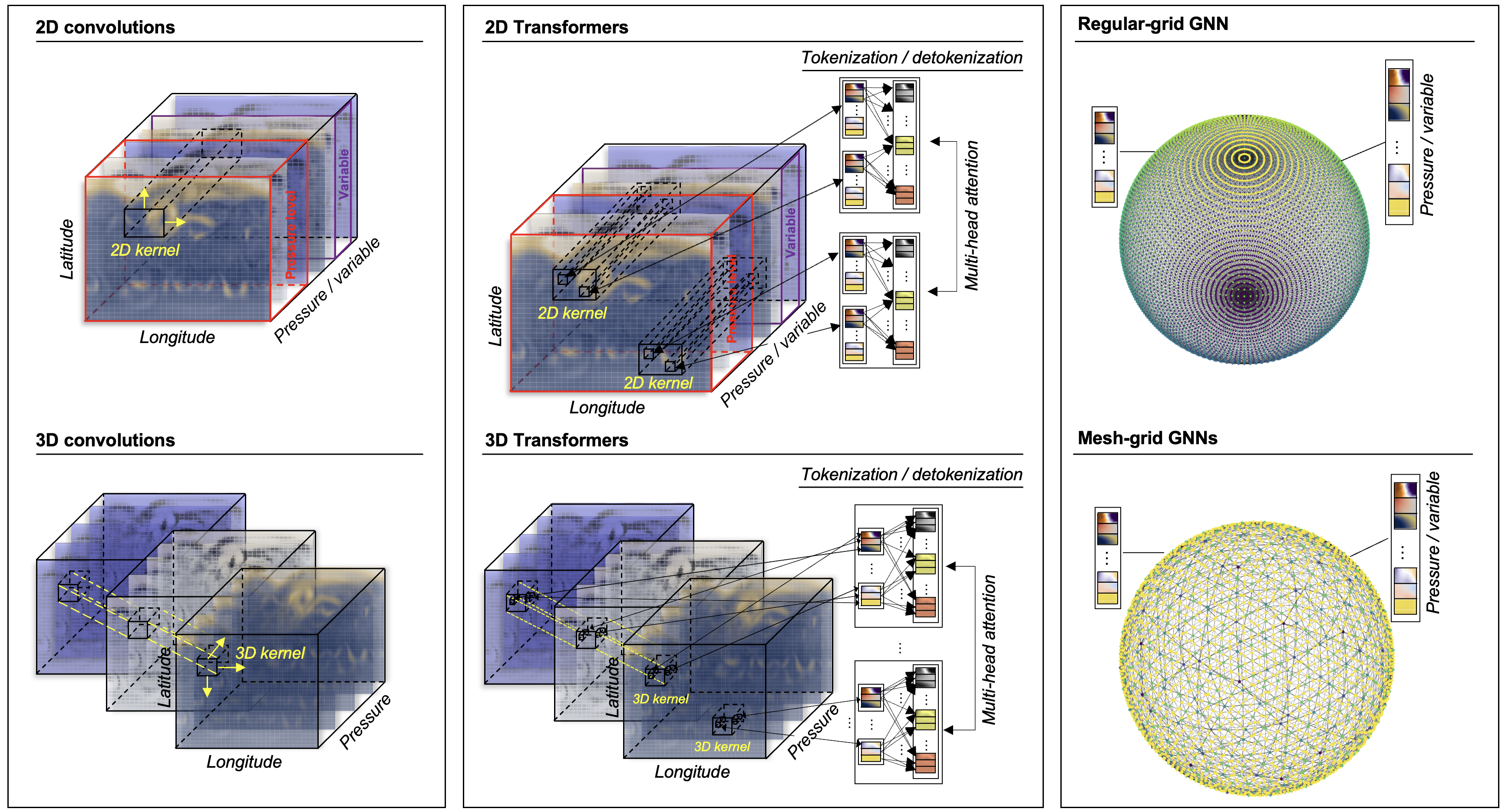}
    \caption{Schematic overview of the architecture families used in this study. 
    Two-dimensional convolutional and transformer models operate on latitude--longitude structure while stacking pressure levels and variables in the channel dimension, whereas three-dimensional variants act directly across latitude, longitude, and pressure and therefore encode explicit vertical coupling. 
    Graph-based models represent the atmosphere on either a spherical grid graph or a spherical mesh while stacking pressure levels and variables in the node features.}
    \label{fig:inductive-biases-figure}
\end{figure}

Given simulator-generated trajectories $\{\mathbf{x}_t\}_{t=1}^T$, emulator learning aims to approximate the one-step time-evolution operator of the simulator. 
We train a neural network $f_{\theta}$ to predict
\begin{equation}
    \hat{\mathbf{x}}_{t+\Delta t}=f_{\theta}(\mathbf{x}_t),
\end{equation}
where $\mathbf{x}_t$ is the atmospheric state and $\Delta t=3$ hours. 
The model parameters are optimized on paired examples $(\mathbf{x}_t,\mathbf{x}_{t+\Delta t})$ by minimizing a prediction loss over the training set,
\begin{equation}
    \theta^\ast
    =
    \arg\min_{\theta}
    \frac{1}{N}\sum_{i=1}^{N}
    \mathcal{L}\!\left(f_{\theta}(\mathbf{x}^{(i)}_t),\mathbf{x}^{(i)}_{t+\Delta t}\right).
\end{equation}

The central question is how architectural inductive bias affects this approximation. 
We compare convolutional, graph-based, and transformer-based emulators that differ in how they represent horizontal structure, vertical coupling, and long-range interaction. 
In particular, 2D models operate on latitude--longitude fields while stacking pressure levels and variables, whereas 3D models operate directly over latitude, longitude, and pressure. 
Graph-based models represent the atmosphere on spherical grid or mesh graphs while stacking pressure levels and variables in node features. 
This design allows us to test whether explicit vertical coupling becomes more important when SSW-like wave dynamics are active. 
A schematic overview of the architecture families is shown in Figure~\ref{fig:inductive-biases-figure}.

All models are trained for the same one-step forecasting task, predicting temperature, zonal wind, and meridional wind jointly over the full grid.
Within each architecture family, we run a fixed-size hyperparameter sweep of 50 configurations and select the model with the lowest validation MAE. 
Model skill is evaluated on the held-out test period using per-variable mean absolute error over the full atmosphere and over the mid-upper stratosphere, defined here as pressure levels with pressures lower than 30 hPa. 
To test whether low pointwise error also corresponds to physically meaningful dynamics, we additionally compute Eliassen--Palm flux diagnostics from the predicted and true fields. 
Further details of the preprocessing, model families, hyperparameter sweeps, and implementation choices are provided in the Supporting Information.

\section{Results}

\subsection{Overall summary}

\begin{table}[t]
  \centering
  \renewcommand{\arraystretch}{1.1}
  \setlength{\tabcolsep}{4pt}
  \caption{\small Best model per architecture on the no-SSW and SSW-enabled Isca datasets. Reported values show mean absolute error (MAE) for temperature, zonal wind, and meridional wind over the full atmosphere and over the stratosphere ($p < 30$ hPa). \best{Dark}, \second{medium}, and \third{light green} text indicate the best, second-best, and third-best values in each column, respectively.}
  \small
  \resizebox{\textwidth}{!}{%
  \begin{tabular}{l ccc ccc c ccc ccc c}
    \toprule
      & \multicolumn{7}{c}{\textbf{No SSW}}
      & \multicolumn{7}{c}{\textbf{With SSW}} \\
      \cmidrule(lr){2-8} \cmidrule(lr){9-15}
      & \multicolumn{3}{c}{\textbf{Full atmosphere}}
      & \multicolumn{3}{c}{\textbf{Stratosphere ($<\!30\,\mathrm{hPa}$)}}
      & \multicolumn{1}{c}{}
      & \multicolumn{3}{c}{\textbf{Full atmosphere}}
      & \multicolumn{3}{c}{\textbf{Stratosphere ($<\!30\,\mathrm{hPa}$)}}
      & \multicolumn{1}{c}{} \\
      \cmidrule(lr){2-4} \cmidrule(lr){5-7}
      \cmidrule(lr){9-11} \cmidrule(lr){12-14}
      \textbf{Architecture}
      & \textbf{\shortstack[c]{Temp \\ $[K]$}}
      & \textbf{\shortstack[c]{Z. wind \\ $[m/s]$}}
      & \textbf{\shortstack[c]{M. wind \\ $[m/s]$}}
      & \textbf{\shortstack[c]{Temp \\ $[K]$}}
      & \textbf{\shortstack[c]{Z. wind \\ $[m/s]$}}
      & \textbf{\shortstack[c]{M. wind \\ $[m/s]$}}
      & \textbf{\shortstack[c]{Num.\\ params}}
      & \textbf{\shortstack[c]{Temp \\ $[K]$}}
      & \textbf{\shortstack[c]{Z. wind \\ $[m/s]$}}
      & \textbf{\shortstack[c]{M. wind \\ $[m/s]$}}
      & \textbf{\shortstack[c]{Temp \\ $[K]$}}
      & \textbf{\shortstack[c]{Z. wind \\ $[m/s]$}}
      & \textbf{\shortstack[c]{M. wind \\ $[m/s]$}}
      & \textbf{\shortstack[c]{Num.\\ params}} \\
    \midrule
    Persistence
      & 0.244 & 0.581 & 0.621 & 0.352 & 0.790 & 0.778 & --
      & 0.235 & 0.608 & 0.678 & 0.316 & 0.818 & 0.863 & -- \\
    Simple CNN2D
      & 0.088 & 0.170 & 0.146 & 0.127 & 0.241 & 0.192 & 154M
      & 0.119 & 0.263 & 0.251 & 0.164 & 0.373 & 0.348 & 157M \\
    CNN2D U\mbox{-}Net
      & 0.080 & 0.173 & 0.164 & 0.119 & 0.252 & 0.226 & 36M
      & 0.123 & 0.284 & 0.288 & 0.176 & 0.419 & 0.418 & 147M \\
    Simple CNN3D
      & 0.081 & 0.151 & 0.122 & 0.114 & 0.208 & 0.154 & \second{6M}
      & 0.084 & 0.190 & 0.195 & 0.117 & 0.277 & 0.282 & \second{6M} \\
    CNN3D U\mbox{-}Net
      & \third{0.060} & \third{0.111} & \second{0.082} & \third{0.083} & \third{0.152} & \second{0.100} & \best{3M}
      & \second{0.070} & \second{0.150} & \second{0.141} & \second{0.095} & \second{0.214} & \second{0.195} & \best{5M} \\
    Trf.\mbox{-}2D GA
      & 0.083 & 0.179 & 0.161 & 0.119 & 0.244 & 0.203 & 679M
      & 0.118 & 0.281 & 0.267 & 0.162 & 0.395 & 0.365 & 214M \\
    Trf.\mbox{-}2D Swin
      & 0.080 & 0.181 & 0.167 & 0.115 & 0.247 & 0.217 & 217M
      & 0.109 & 0.267 & 0.267 & 0.154 & 0.381 & 0.375 & 259M \\
    Trf.\mbox{-}3D GA
      & \second{0.046} & \best{0.090} & \best{0.075} & \second{0.063} & \best{0.122} & \best{0.093} & 101M
      & \best{0.055} & \best{0.126} & \best{0.129} & \best{0.078} & \best{0.181} & \best{0.187} & 134M \\
    Trf.\mbox{-}3D Swin
      & \best{0.044} & \second{0.103} & \third{0.101} & \best{0.060} & \second{0.132} & \third{0.121} & 82M
      & \third{0.074} & \third{0.167} & \third{0.171} & \third{0.104} & \third{0.244} & \third{0.246} & 82M \\
    Grid MPNN
      & 0.096 & 0.213 & 0.206 & 0.145 & 0.308 & 0.289 & \third{9M}
      & 0.140 & 0.349 & 0.357 & 0.200 & 0.508 & 0.511 & \third{8M} \\
    Mesh MPNN
      & 0.094 & 0.205 & 0.191 & 0.143 & 0.300 & 0.271 & 31M
      & 0.129 & 0.295 & 0.278 & 0.180 & 0.428 & 0.398 & 21M \\
    \bottomrule
  \end{tabular}%
  }
  \label{tab:main_results}
\end{table}

Table~\ref{tab:main_results} summarizes the best tuned model within each architecture family on the no-SSW and SSW-enabled datasets. 
Across both regimes, but most clearly in the SSW-enabled case and in the stratosphere-focused metrics, explicitly three-dimensional models perform best, especially the 3D U-Net and the 3D global transformer. 
By contrast, the 2D and graph-based models are generally less competitive in the stratosphere-focused metrics, despite in some cases having substantially more parameters. 
The model sizes were selected by the validation-based hyperparameter tuning procedure, rather than imposed manually.

\subsection{Performance differences widen in the SSW-enabled regime}
\begin{figure}[t]
    \centering
    \includegraphics[width=\linewidth]{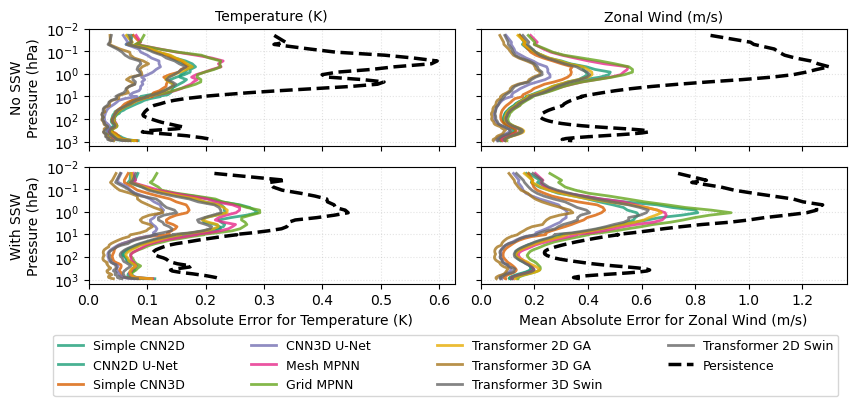}
    \caption{\small 
    Mean absolute error as a function of pressure level for temperature and zonal wind. Colored lines denote learned architectures and the dashed black line denotes persistence.}
    \label{fig:mae_error_structure}
\end{figure}

Figure~\ref{fig:mae_error_structure} shows mean absolute error as a function of pressure level for temperature and zonal wind in the no-SSW and SSW-enabled regimes. 
The learned models are clearly separated from the persistence baseline, indicating that all architectures learn meaningful predictive structure beyond simply carrying the previous state forward. 
The persistence errors are pronounced in both the no-SSW and with-SSW regimes, despite the lack of zonal-mean variability in the no-SSW case as seen in Figure~\ref{fig:ssw_examples}. 
The reason the no-SSW configuration has a large persistence error aswell is that small displacements in the sharp wind structure generate large grid-point errors.

In the no-SSW case, the learned models themselves remain relatively clustered, with only moderate architecture-dependent differences in their vertical error profiles. 
By contrast, in the SSW-enabled regime, the spread between learned architectures increases markedly in the upper troposphere and stratosphere. 
This is particularly clear for zonal wind, where the strongest vertically structured models maintain substantially lower error than the weaker 2D and graph-based models.
Hence, the SSW-enabled regime does not simply increase error uniformly across all models; rather, it amplifies the performance differences between model classes.
This suggests that vertically structured architectures provide not only a lower domain-averaged MAE, but also a more accurate representation of the complex stratospheric dynamics. 

\subsection{EP-flux diagnostics}
\label{sec:epflux_results}
Accurate pointwise prediction of these variables does not necessarily imply that the learned emulator preserves the physical relationships between them. 
This is particularly important for EP-flux diagnostics, which depend on the eddy covariance terms $\overline{u'v'}$ and $\overline{v'T'}$. 
Thus, a model may reproduce the individual fields well while still distorting the wave-driving structure that controls wave-mean-flow interaction.

\begin{figure}[t]
    \centering
    \includegraphics[width=\linewidth]{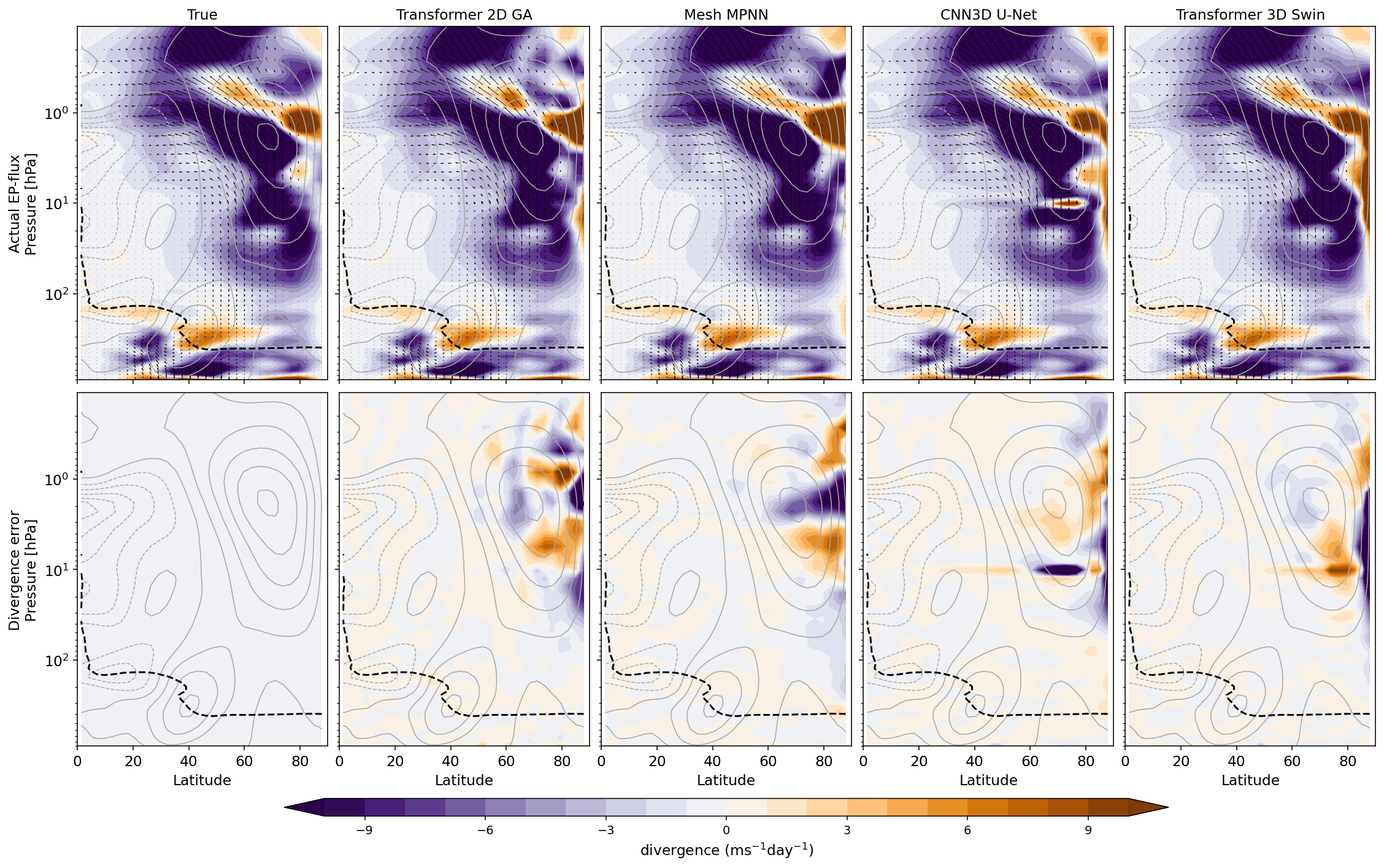}
    \caption{\small EP-flux diagnostics for selected architectures averaged over the 10 days preceding a held-out SSW event.
    The upper row shows EP-flux vectors overlaid on EP-flux divergence for the true simulator trajectory and model predictions.
    The lower row shows divergence error, defined as true minus predicted divergence.
    The dashed contour marks the dynamical tropopause, defined here as the zonal- and time-mean contour of absolute potential vorticity.}   
    \label{fig:epfluxselected}
\end{figure}

Figure~\ref{fig:epfluxselected} shows EP-flux diagnostics averaged over the 10 days preceding the vortex-weakening event shown in Figure~\ref{fig:ssw_examples} (c).
The event is chosen as a clear example of vortex weakening, identified from the zonal-mean zonal wind at 60$^\circ$N and 10 hPa, and occurs more than 125 days after the training, validation, and test periods. 
This pre-event period is used because anomalous upward and poleward wave activity is expected before the vortex weakening~\cite{baldwin2021sswreview}.
For each model, we compute EP fluxes from the predictions and compare them with the corresponding diagnostic computed from the true simulator trajectory. 
Rather than evaluating an autoregressive rollout, the EP-flux diagnostics are averaged over the $t+1$ outputs of 80 independent one-step predictions during the 10-day pre-event period. 
This avoids attributing diagnostic errors to accumulated trajectory drift, where small state errors can move the predicted flow away from the event being diagnosed. 
Instead, the errors shown in these diagnostics provide a short-lead-time test of physical consistency, assessing whether the emulator preserves the wind--temperature covariance relationships needed to reproduce the EP-flux structure central to stratospheric wave-mean-flow interaction.
The upper row shows the EP-flux vectors and EP-flux divergence, while the lower row shows the divergence error, defined as true minus predicted divergence.
The upper panels show that the emulators preserve the EP-flux structure at a broad level where the predicted divergence fields place the largest positive and negative regions in approximately the right parts of the latitude-pressure domain. 
The largest errors occur in the high-latitude stratosphere (1-10 hPa), especially near the region where the true fields show strong pre-SSW EP-flux divergence.
These errors are coherent regions of over- and under-predicted divergence, indicating that the models misplace or misestimate part of the wave forcing.
By contrast, tropospheric errors are generally smaller, even though the true tropospheric EP-flux divergence is also large. 
This suggests that the stratospheric EP-flux diagnostic is not harder merely because the diagnostic amplitude is larger, but because the relevant covariance structure is more difficult to reproduce during SSW development.

\begin{figure}[t]
    \centering
    \includegraphics[width=\linewidth]{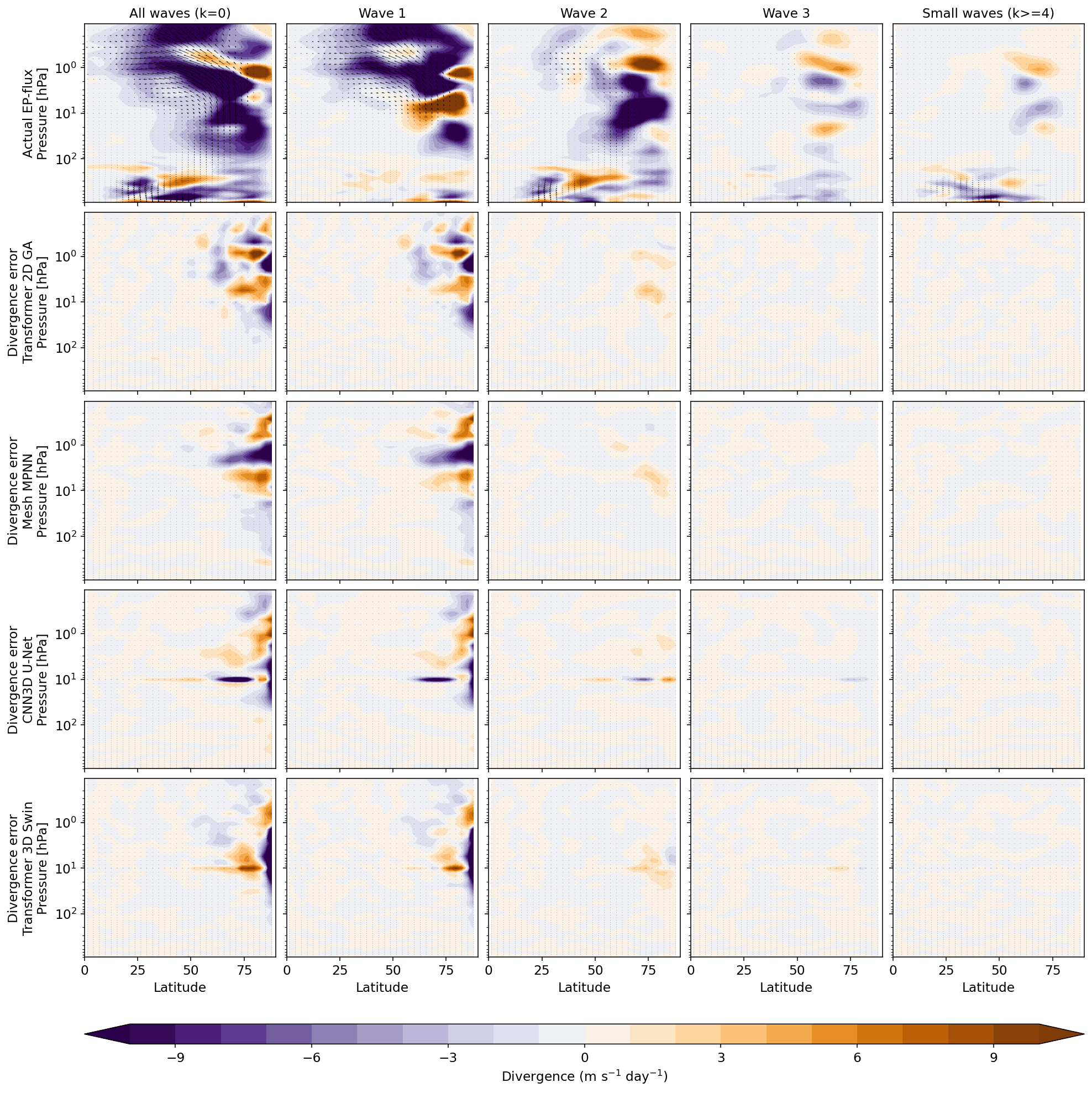}
    \caption{\small Zonal-wavenumber decomposition of EP-flux divergence during the 10 days preceding a held-out SSW event. 
    Columns show how the EP-flux divergence is split by wave scale: all waves together, wave $1$, wave $2$, wave $3$, and the remaining smaller-scale waves.
    The top row shows the true simulator diagnostic, while lower rows show divergence errors for selected emulators, defined as true minus predicted divergence. 
    The largest coherent errors are concentrated in the wave-1 component.}
    \label{fig:epflux_wave_split}
\end{figure}

We decompose the EP fluxes by scale by recomputing them using only zonal wave $1$, wave $2$, and wave $3$ separately using a spatial Fourier-Transform over longitude~\cite{jucker_2021} and define residual smaller-scale contributions as the full diagnostic minus the contributions from waves $1$-$3$.
Figure~\ref{fig:epflux_wave_split} shows this decomposition for the true simulation and for selected emulators. 
In the Isca simulation, the pre-SSW stratospheric EP-flux divergence is dominated by waves $1$ and $2$. 
However, the emulator error is not distributed in proportion to the EP-flux divergence strength across wavenumbers, but is especially large for the wave-$1$ component in the high-latitude stratosphere.
The contrast between waves 1 and 2 is particularly intriguing given that the main source of tropospheric planetary-scale variability comes from our imposed wave-2 heating.
Therefore, any significant wave-1 component of the EP-flux must come from internal atmospheric interactions, which are likely to be more difficult for the emulators to predict. 
However, it is not simply the case that wave 1 is intrinsically hard to predict - the wave-1 tropospheric EP-flux anomalies are predicted with low error by the emulators, as is the total EP-flux divergence from all waves. 

The contrast we observe between waves 1 and 2 in the EP-flux divergence error is consistent with previous research. 
\citet{Lindgren_2020} showed that stratospheric wave-wave interactions can transfer energy between wavenumbers, producing SSW structures in a wave number different from that of the tropospheric forcing. 
Furthermore \citet{Lindgren_2020}, showed that selectively removing wave-wave interactions in the upper stratosphere significantly alters SSW frequencies.
The wave-1 stratospheric EP-flux signal in our SSW-enabled simulation, despite imposed wave-2 tropospheric forcing, could be expected to arise predominantly from in-stratosphere wave-wave interactions.
Seeing that the emulators reproduce tropospheric wave-1 EP-flux structure but systematically misplace stratospheric wave-1 EP-flux divergence therefore suggests a specific failure mode where architectures capture the wave-mean-flow response to the imposed wave-2 forcing but struggle with the nonlinear stratospheric processes responsible for the wave-1 signal. 
Because the training period already contains disturbed vortex states, including both split-like and displacement-like configurations, this error is unlikely to be resolved simply by exposing the models to more event morphologies. 
Instead, the diagnostic points to a need for architectures that better preserve the physical covariance structure underlying wave activity.

A narrow error band near 10 hPa appears for 3D architectures in Figures~\ref{fig:epfluxselected}--\ref{fig:epflux_wave_split}. 
This discretization-related artifact arises from including an exact 10 hPa diagnostic level close to an existing 10.24 hPa model level. 
We retain it as a useful emulator-design example: physically derived diagnostics can expose artifacts caused by seemingly minor preprocessing choices, even when pointwise forecast errors appear unaffected. 
The artifact is removed to proof this point and discussed more in the Supporting Information.

\section{Limitations \& future research}
Our results show that explicit vertical coupling is particularly important when stratospheric dynamics are active. However, the EP-flux diagnostics highlight the need for more physically consistent emulators.
Firstly, physically informed objectives could combine EP-flux or tendency-consistency terms with vortex-geometry measures such as orientation, ellipticity, and centroid, which are particularly relevant for wave-1 variability~\cite{Daniel_2011, Seviour_2013, Waugh_1997}.
Secondly, because this study deliberately focuses on one-step predictions to isolate architectural inductive biases, multi-step and event-conditioned training objectives are a natural next step.
Third, the relatively good qualitative behaviour of the mesh-based GNN in some diagnostics suggests that spherical geometry may matter even when its aggregate MAE remains less competitive, especially if combined with explicit vertical structure.
Finally, the models are trained using a relatively short model timeseries which was possible since the Isca configuration was constantly disrupting the vortex in the simulation.
However, one focus for future work would be to study the various types and precursors of different SSWs, and to look specifically at how these different classes of SSWs are handled by the various architectures~\cite{baldwin2021sswreview}. 

\section{Conclusion}
We benchmarked machine-learning emulator architectures on idealised Isca simulations with and without SSW-like dynamics. 
While model differences are relatively modest in the no-SSW regime, they widen substantially in the SSW-enabled regime, where architectures with explicit three-dimensional structure perform best. 
At the same time, mechanism diagnostics show that low forecast error alone is not sufficient: even strong models can still fail to fully preserve the relevant wave-driving structure during active events. 
In particular, the largest coherent diagnostic errors occur in the high-latitude stratosphere and are concentrated in the long-wave divergence structure. 
Together, these results show that robust atmospheric emulation requires both appropriate architectural inductive biases, especially explicit vertical coupling, and evaluation metrics that test physically consistent behaviour beyond grid-point error.

\section*{Code and Data Availability}
The fork of \textsc{Isca} and the configuration files used to generate the no-SSW and SSW-enabled simulations are available in the \href{https://github.com/oskarbohnlassen/isca_emulator_datageneration/tree/paper-datageneration}{Isca data-generation repository}. 
The machine-learning and analysis code is available in the \href{https://github.com/oskarbohnlassen/isca_emulation}{Isca emulation repository}. 
This repository contains scripts for preprocessing raw \textsc{Isca} output, constructing train-validation-test datasets, training the emulator models, evaluating predictions, computing Eliassen--Palm flux diagnostics, and reproducing the figures in this paper. 
The corresponding experiment logs, final training runs, validation metrics, model configurations, and selected model artifacts are available in the accompanying \href{https://api.wandb.ai/links/obola-the-technical-university-of-denmark/snpv7zhn}{Weights \& Biases report}. 
The code for computing EP fluxes was adapted from \href{https://zenodo.org/records/17093152}{\texttt{aostools}}~\cite{jucker_2021}. 
Together, these repositories provide the code required to regenerate the simulation data products and reproduce the experiments reported here.

\section*{Acknowledgments}
This work is supported by Novo Nordisk Foundation grant NNF23OC0085356 and Pioneer Centre for AI, DNRF grant number P1.

\bibliographystyle{unsrtnat}
\bibliography{references} 

\clearpage

\appendix

\section{Detailed dynamics of Isca and elaboration of Eliassen--Palm fluxes}
\label{si:isca}

Isca is a modular atmospheric modelling framework designed to support controlled experiments across a hierarchy of model complexities, from idealised dry dynamical cores to more comprehensive GCM configurations~\cite{vallis2018isca}. 
This makes it particularly suitable for studying how specific physical processes affect atmospheric dynamics while minimizing confounding effects from unrelated model components.

At each time $t$, the simulator state is represented by a collection of atmospheric variables defined on a spherical latitude-longitude-pressure grid,
\begin{equation}
    \mathbf{x}_{t} \in \mathbb{R}^{V \times L_{\phi} \times L_{\lambda} \times L_{p}},
\end{equation}
where $V$ denotes the number of atmospheric variables, $L_{\phi}$ the number of latitudes, $L_{\lambda}$ the number of longitudes, and $L_{p}$ the number of vertical pressure levels. 
In our experiments, these variables include temperature $T$, zonal wind $u$, and meridional wind $v$, evaluated on a regular T42 grid corresponding to $64 \times 128$ latitude-longitude points and 60 vertical levels. 
The simulator advances the atmospheric state in discrete time by numerically integrating the underlying governing equations, producing,
\begin{equation}
    \mathbf{x}_{t+\Delta t} = \mathcal{M}(\mathbf{x}_t),
\end{equation}
where $\mathcal{M}$ denotes the one-step time-evolution operator induced by the chosen \textsc{Isca} configuration.

In this work, we use a dry spectral dynamical-core configuration based on \cite{mudhar2024modeldependency}. 
This setup follows the Held-Suarez style idealization in the troposphere, where temperature is relaxed toward a prescribed radiative-equilibrium profile and low-level winds are damped through Rayleigh friction \cite{held1994dynamicalcores}. 
Abstractly, the continuous-time evolution may be written as,
\begin{equation}
\label{eq:dry_core_forcing}
    \frac{\partial \mathbf{x}}{\partial t} = \mathcal{D}\!\left(\mathbf{x}; T_{\mathrm{eq}}\right) + \mathbf{F}_{\mathrm{wave}},
\end{equation}
where $\mathcal{D}(\cdot)$ denotes the dry dynamical-core tendencies, including advection, pressure-gradient forces, Newtonian relaxation toward the equilibrium temperature profile $T_{\mathrm{eq}}$, and frictional damping, while $\mathbf{F}_{\mathrm{wave}}$ denotes additional imposed forcing terms which for the SSW-disabled case is set to $0$.

The equilibrium temperature profile is piecewise defined by tropospheric and stratospheric components,
\begin{equation}
    T_{\mathrm{eq}}(p,\phi)=
    \begin{cases}
        T_{\mathrm{eq}}^{\mathrm{trop}}(p,\phi), & p \ge p_T,\\
        T_{\mathrm{eq}}^{\mathrm{strat}}(p,\phi), & p < p_T,
    \end{cases}
\end{equation}
where $p$ denotes pressure, $\phi$ latitude, and $p_T$ the tropopause transition pressure. 
The tropospheric part follows the standard Held-Suarez relaxation profile, while the stratospheric part follows a modified Polvani-Kushner formulation~\cite{polvani2002stratosphere}.
A key control parameter is the stratospheric lapse-rate parameter $\gamma$, which controls the vertical temperature structure of the stratospheric equilibrium profile and thereby influences the mean strength and variability of the winter polar vortex. The full functional form of $T_{\mathrm{eq}}^{\mathrm{strat}}$ is given in~\cite{mudhar2024modeldependency}.

To generate the more dynamically challenging SSW-enabled regime used in this paper, we activate the additional prescribed heating perturbation following \cite{mudhar2024modeldependency} and \cite{Lindgren_2018}.
This perturbation enters equation~\eqref{eq:dry_core_forcing} as the imposed tendency $\mathbf{F}_{\mathrm{wave}}$ and increases stratospheric wave activity and variability.
The perturbation takes the form of a zonal wave-$2$ heating pattern in longitude, centred at approximately 45$^\circ$N and 400 hPa. 
This acts to generate stationary planetary-scale Rossby waves, which propagate upward into the stratosphere and disturb the winter polar vortex. 
Such a forcing is required in this model because, in the real atmosphere, planetary waves are generated in part by flow over topography, which we omit in our idealised setup. 
However, prescribing the planetary-wave forcing in this way allows us to control the magnitude of the stratospheric wave forcing and to create two parallel datasets with and without strong stratospheric variability.

This distinction is visualized in the main text as Figure~1, which shows the zonal-mean zonal wind at $60^\circ$N and 10 hPa over time for the two configurations used in this study as well as absolute vorticity for selected time-indices. 
The configuration without additional forcing exhibits a stronger and more stable polar vortex, whereas the forced configuration displays larger temporal fluctuations and intermittent vortex weakening events characteristic of an SSW-enabled regime. 
These two settings form the basis of our machine-learning benchmark: a relatively simple regime without pronounced stratospheric disruptions, and a more challenging regime in which vertically coupled wave dynamics are active.

The full planetary configuration is intentionally idealised: there is no land surface, no topography, and no moisture. 
Although simplified, this setup is not arbitrary. 
It is designed to isolate large-scale dry dynamical mechanisms relevant for stratosphere-troposphere coupling.
Although the wind fields are not identical, the idealised model produces a qualitatively similar large-scale structure, including a strong Northern Hemisphere winter polar vortex. 

\subsection{Wave-mean-flow interactions \& Eliassen-Palm fluxes}

To understand sudden stratospheric warmings, it is not sufficient to consider only the zonal-mean flow. 
One must also account for waves and eddies in the flow, since they can transport momentum and heat and thereby influence the zonal-mean circulation.
In atmospheric science, this is typically done by decomposing fields such as temperature $T$, zonal wind $u$, and meridional wind $v$ into a zonal mean and a deviation from that mean. 
At a fixed latitude, pressure level, and time, these deviations describe how the fields vary around the globe in longitude. 
We refer to these longitude-dependent deviations as eddy fields, which include the planetary-scale waves central to SSW dynamics.

Each field can be decomposed into a mean and a deviation from that mean, for example for zonal wind,
\begin{equation}
    u(\lambda,\phi,p,t)=\overline{u}(\phi,p,t)+u'(\lambda,\phi,p,t),
\end{equation}
and similarly for $v$ and $T$. 
The zonal mean $\overline{u}$ represents the background flow, while the deviation $u'$ captures the longitudinal wave structure. 
In this setting, planetary or Rossby waves correspond to large-scale wave-like patterns in these deviations, that is, coherent alternating structures in wind and temperature around a latitude circle~\cite{Vallis_2017}.

These waves are present in both \textsc{Isca} configurations, but the additional imposed heating perturbation strengthens them and makes their interaction with the winter stratosphere more pronounced. 
This matters because the waves do not simply create local anomalies in $T$, $u$, and $v$; they also transport momentum and heat through the atmosphere. 
For example, if poleward meridional wind anomalies $v'$ tend to occur at the same longitudes as positive zonal-wind anomalies $u'$, then the wave field transports zonal momentum poleward. 
Likewise, if poleward meridional wind anomalies tend to co-occur with warm temperature anomalies $T'$, then the wave field transports heat. 
These effects are summarized by the eddy covariances,
\begin{equation}
    \overline{u'v'}
    \qquad\text{and}\qquad
    \overline{v'T'},
\end{equation}
which are zonal averages of products of the longitude-dependent deviations from the zonal mean, evaluated at fixed time, latitude, and pressure. 
Here, $\overline{u'v'}$ measures the covariance between zonal-wind and meridional-wind anomalies, while $\overline{v'T'}$ measures the covariance between meridional-wind and temperature anomalies.

Wave-mean-flow interaction refers to the fact that these waves propagate through the atmosphere and can transfer momentum to the zonal-mean flow when they are absorbed or dissipated. 
In the winter hemisphere, much of the wave activity remains confined to the troposphere, but some propagates upward into the stratosphere. 
When this upward-propagating wave activity becomes strong enough, it can disturb and weaken the polar vortex. 
This is one of the central dynamical ingredients of SSW-like events, and it is therefore highly relevant for the present benchmark.

To analyze the process of wave-mean-flow interaction, we use Eliassen-Palm (EP) flux diagnostics~\cite{andrews_1987}. 
The first part of the diagnostics is the EP-flux arrow, defined as a vector field in the latitude-pressure space,
\begin{equation}
    \mathbf{F} = (F_{\phi},F_{p}),
\end{equation}
where $F_{\phi}$ is the component in the meridional direction and $F_{p}$ is the component in the vertical pressure direction. 
In the present work we make use of the quasi-geostrophic (QG) approximation of the EP-fluxes, meaning that the components are proportional to the eddy covariances of both momentum and heat,
\begin{equation}
    F_{\phi} \propto -\,\overline{u'v'}
    \qquad
    F_{p} \propto \overline{v'T'}.
\end{equation}
The full implementation of these QG EP-fluxes further accounts for spherical geometry and static stability (see e.g., \cite{Martineau_2018}), and can be found in the attached code.
The direction of the EP flux arrows indicate the relative importance of the eddy fluxes of heat and momentum. In the case where the eddy dynamics is purely Rossby waves, the EP-flux arrows indicate how wave activity propagates through latitude-pressure space. 
An upward arrow indicates upward propagation, while a poleward arrow indicates propagation toward high latitudes \cite{EliassenPalmCrossSectionsfortheTroposphere}. 
In our SSW-enabled \textsc{Isca} configuration, stronger upward and poleward propagation into the winter stratosphere is expected, particularly toward the polar vortex region.

The second part of the diagnostic is the EP-flux divergence term,
\begin{equation}
    \nabla \cdot \mathbf{F}
    =
    \frac{1}{a\cos{\phi}}\frac{\partial (F_{\phi}\cos{\phi})}{\partial \phi}
    +
    \frac{\partial F_{p}}{\partial p},
\end{equation}
where $a$ is the Earth's radius. The divergence of $\mathbf{F}$ measures how the EP-flux arrows change across latitude-pressure space. 
Mathematically, the divergence is the sum of the spatial derivatives of the meridional and vertical components of the vector field, so it compares how much EP-flux enters a small latitude-pressure region with how much leaves it. 
If the inflow and outflow are locally balanced, the divergence is zero; if they are not, the divergence is nonzero and indicates a net accumulation or net export of wave activity. 
This is physically useful because regions where wave activity accumulates or is removed are precisely the regions where it interacts most strongly with the mean flow and can therefore influence the zonal-mean zonal wind.

In the context of SSWs, one typically expects enhanced upward and poleward EP-flux into the winter stratosphere together with strong wave forcing in the polar-vortex region. 
Hence, from a physical perspective, accurately predicting the atmospheric state prior to and during an SSW is closely related to reproducing the correct wave-mean-flow interactions. 
This makes EP-flux diagnostics a useful mechanism-aware complement to standard error metrics. 
In particular, they test whether a model predicts not only the correct pointwise values of $T$, $u$, and $v$, but also the covariance structure between them that governs realistic wave propagation and mean-flow forcing.

\noindent\section{Machine-learning emulator architectures used in the study}
\label{si:ml_architectures}

Given simulator-generated trajectories $\{\mathbf{x}_t\}_{t=1}^T$, the goal of emulator learning is to approximate the one-step time-evolution operator of the simulator. 
That is, we seek a parametric function $f_{\theta}$ such that,
\begin{equation}
    \hat{\mathbf{x}}_{t+\Delta t} = f_{\theta}(\mathbf{x}_t),
\end{equation}
where $\mathbf{x}_t$ denotes the atmospheric state at time $t$, $\hat{\mathbf{x}}_{t+\Delta t}$ the predicted state at the next time step, and $\theta$ the learnable model parameters. 
In our setting, $f_{\theta}$ is implemented by a neural network trained on paired input-target examples $(\mathbf{x}_t,\mathbf{x}_{t+\Delta t})$ extracted from the \textsc{Isca} simulations.

Training is performed by optimizing the model parameters to minimize the average prediction loss over the simulator-generated training set,
\begin{equation}
    \theta^\ast
    =
    \arg\min_{\theta}
    \frac{1}{N}\sum_{i=1}^{N}
    \mathcal{L}\!\left(f_{\theta}(\mathbf{x}^{(i)}_t),\mathbf{x}^{(i)}_{t+\Delta t}\right),
\end{equation}
where $\mathcal{L}$ is a prediction loss, $N$ is the number of training samples, and the optimization is performed with gradient-based methods. 
Thus, the emulator does not explicitly solve the governing physical equations, but instead learns a statistical approximation of the simulator time-step map from data.

A central question in this work is how the choice of neural-network architecture affects the quality of this approximation. 
Although all models ultimately map tensors to tensors, different architecture families encode different assumptions about spatial locality, vertical coupling, long-range interaction, and the symmetry structure of the data. 
These architectural assumptions define the model's \emph{inductive bias}, and may become particularly important when the underlying atmospheric dynamics are more difficult to emulate.

The models differ primarily in how they represent the atmospheric state: 2D architectures operate on the horizontal latitude-longitude grid while treating pressure levels as channels, 3D architectures operate directly over latitude, longitude, and pressure, and graph-based models replace the regular grid with either a spherical grid graph or a spherical mesh.

\subsection{Convolutional neural networks}
\label{si:cnns}
Convolutional neural networks (CNNs) are designed for data defined on regular grids, such as images, videos, or geophysical fields represented on regular latitude-longitude grids. 
Their core idea is to learn small local filters that are applied repeatedly across the domain using the same weights. 
In this way, the model does not learn a separate set of parameters for every grid point, but instead learns spatial patterns that can appear anywhere in the field. 
This makes CNNs particularly well suited for atmospheric data, where nearby locations are often strongly related and where similar spatial structures may occur in different regions.

Mathematically, a convolutional layer computes each output value as a weighted sum of values in a local neighborhood of the input,
\begin{equation}
    \mathbf{Y}_{i,j,c_{\mathrm{out}}}
    =
    \sum_{a,b,c_{\mathrm{in}}}
    \mathbf{K}_{a,b,c_{\mathrm{in}},c_{\mathrm{out}}}
    \mathbf{X}_{i+a,j+b,c_{\mathrm{in}}}
    + b_{c_{\mathrm{out}}},
\end{equation}
where $\mathbf{X}$ is the input tensor, $\mathbf{Y}$ is the output tensor, $\mathbf{K}$ is a learnable kernel, $b_{c_{\mathrm{out}}}$ is a bias term, $(i,j)$ indexes the spatial grid location, $c_{\mathrm{in}}$ and $c_{\mathrm{out}}$ index input and output channels, and $(a,b)$ indexes offsets within the local patch of grid points.
For a 3D convolution, the same expression is extended by replacing the horizontal location $(i,j)$ with a three-dimensional location $(i,j,z)$ and the horizontal offsets $(a,b)$ with three-dimensional offsets $(a,b,d)$, so that the sum is also taken over neighboring pressure levels.
Intuitively, for each grid location $(i,j)$, the layer looks only at a small surrounding patch of the input, combines these nearby values with learned weights, and produces a new representation at that location. 
The same kernel is used across all grid locations, which is the source of both the efficiency and the spatial inductive bias of CNNs.

In this benchmark, the 2D CNN treats latitude and longitude as the spatial grid and stacks the atmospheric variables, and also vertical levels, along the channel dimension. 
This is computationally efficient, but vertical interactions are then represented only indirectly through channel mixing. 
By contrast, a 3D CNN treats the atmosphere as a genuine three-dimensional field over latitude, longitude, and pressure, so that the convolutional kernels act directly across vertical as well as horizontal neighborhoods. 
This gives the model a stronger inductive bias toward local vertical coupling, at the cost of higher computational and memory requirements.

A related methodological issue is that standard convolutions are approximately translation equivariant. 
This is useful for parameter sharing, but it assumes that the same local pattern should be processed similarly at different locations. 
For atmospheric fields, this assumption is only partly valid, since the dynamics depend on latitude, pressure level, and grid geometry. 
A common relaxation is to augment the prognostic state with fixed positional features such as latitude, longitude, pressure-level indicators, or sinusoidal coordinate encodings. 
This retains convolutional weight sharing while allowing the learned mapping to depend explicitly on position.

Padding defines how the field is extended beyond the computational domain before applying the convolution. 
For longitude, periodicity motivates circular padding, whereas latitudinal and vertical boundaries require additional modelling choices, such as zero, replicated, reflected, or geometry-aware padding.

A particularly important convolutional architecture is the U-Net, which combines local convolutions with an encoder-decoder structure and skip connections \cite{ronneberger2015unet}. 
By progressively coarsening the spatial resolution and then reconstructing it, U-Nets can capture both fine-scale and larger-scale structure, making them well suited for dense prediction tasks on atmospheric fields.
Hence, U-Net-based architectures have been widely used as backbones for data-driven weather and climate prediction, including deep convolutional models applied to gridded global atmospheric fields~\cite{weyn_2020, weyn_2021}.

\subsection{Graph neural networks}
\label{si:gnns}
Graph neural networks (GNNs) extend neural-network computation from regular grids to data represented as a graph. 
A graph consists of nodes and edges: the nodes represent locations or entities, while the edges specify which of these are allowed to exchange information. 
In contrast to CNNs, which assume a fixed rectangular grid and local stencil, GNNs allow the connectivity structure itself to be chosen by the model designer. 
This makes them attractive for atmospheric problems, where one may wish to represent interactions not only on the native latitude-longitude grid, but also on alternative meshes or multi-resolution spatial discretizations. 
They are also appealing when working with data naturally defined on a sphere, since a graph or mesh representation can avoid some of the geometric distortions introduced by regular latitude-longitude grids, especially toward the poles.

In a GNN, each node carries a feature vector describing the local state, and the network updates these node features by repeatedly exchanging information along the graph edges. 
This process is called \emph{message passing}: at each layer, a node receives information from its neighbors, combines these incoming messages, and uses them to update its own representation. 
A generic message-passing layer can be written as,
\begin{equation}
    \mathbf{h}^{(\ell+1)}_i
    =
    \phi^{(\ell)}
    \Bigl(
        \mathbf{h}^{(\ell)}_i,
        \mathrm{AGG}_{j \in \mathcal{N}(i)}
        \psi^{(\ell)}(\mathbf{h}^{(\ell)}_i,\mathbf{h}^{(\ell)}_j,\mathbf{e}_{ij})
    \Bigr),
\end{equation}
where $\mathbf{h}^{(\ell)}_i$ denotes the feature vector of node $i$ at layer $\ell$, $\mathcal{N}(i)$ its set of neighboring nodes, $\mathbf{e}_{ij}$ optional edge features, $\psi^{(\ell)}$ a function that computes messages between connected nodes, $\mathrm{AGG}$ an aggregation operator such as summation, and $\phi^{(\ell)}$ an update function. 
Intuitively, this means that each node updates its state by combining its current information with information received from nearby connected nodes.

For atmospheric emulation, the node features correspond to physical variables such as temperature and wind, edge features can encode geometric information such as relative distance and direction, and the graph structure specifies which spatial locations interact directly.
Nodes may be placed directly on the latitude-longitude grid, or on a mesh-based discretization that provides more flexible or multi-scale connectivity.
In both cases, we account for the spherical geometry of the domain by treating longitude as periodic and by using spherical connectivity near the poles, rather than treating the latitude-longitude grid as a flat rectangular image.

Many modern weather-oriented graph architectures use an encoder-processor-decoder structure: an encoder first maps the raw atmospheric variables to latent node representations, a graph-based processor performs several rounds of message passing, and a decoder maps the final latent features back to predicted physical variables. 
One prominent example is GraphCast~\cite{lam2023graphcast}, which uses message passing on a multi-scale spherical mesh for medium-range weather prediction; another related example is the graph-based weather forecasting model of Keisler~\cite{keisler2022forecastingglobalweathergraph}, which also represents global atmospheric fields using GNNs.
Compared with CNNs, GNNs offer greater flexibility in how spatial interactions are represented and can naturally accommodate non-regular connectivity patterns, although they do not exploit regular-grid structure as directly.

A practical issue when applying GNNs to latitude-longitude fields is that the graph connectivity alone does not encode the orientation or spherical geometry of neighboring nodes. 
For example, a message-passing layer can aggregate information from connected nodes, but without positional or edge information it cannot distinguish whether a neighbor lies to the east, west, north, or south, nor can it account for longitude periodicity or polar geometry. 
We therefore augment the graph representation with geometric information. 
For regular-grid graphs, nodes are assigned trigonometric latitude-longitude features, and edges are assigned relative displacement and distance features. 
Longitude is treated as periodic, and pole-crossing neighbors are handled using spherical wrapping. 
For mesh-based graphs, nodes are represented by their Cartesian coordinates on the unit sphere together with trigonometric latitude-longitude features, while edge attributes encode either global Cartesian displacement and great-circle distance or the sender position expressed in a receiver-local frame. 
These features make the message-passing operation direction- and geometry-aware, while preserving the flexibility of the graph representation.

\subsection{Transformers}
\label{si:transformers}
Transformers operate on data represented as a sequence of \emph{tokens} rather than directly on a grid \cite{vaswani2017attention}. 
For atmospheric fields, a token is a learned representation of a local patch of the input state. 
More concretely, the input tensor is divided into non-overlapping patches, and each patch is mapped to a latent vector of fixed dimension. 
If $\mathbf{x}_t$ denotes the input state, this tokenization step can be written abstractly as
\begin{equation}
    \mathbf{H}^{(0)} = \mathrm{Tokenize}(\mathbf{x}_t) + \mathbf{P},
\end{equation}
where $\mathbf{H}^{(0)} \in \mathbb{R}^{N \times d}$ is the resulting sequence of $N$ tokens with embedding dimension $d$, and $\mathbf{P}$ denotes positional information added to retain the spatial arrangement of the original field. 
Intuitively, each token summarizes the information in one local patch of the atmosphere, while the positional encoding tells the model where that patch came from.

In a 2D transformer, the atmosphere is represented on a latitude-longitude grid with variables and vertical levels stacked along the channel dimension, so each token corresponds to a spatial patch in latitude and longitude together with all channels at those locations. 
In a 3D transformer, the tokenization instead acts on latitude, longitude, and pressure jointly, so each token represents a three-dimensional atmospheric volume. 
Thus, analogous to the distinction between 2D and 3D convolutions, the choice of tokenization determines whether vertical structure is handled implicitly through channels or explicitly as part of the spatial input representation.

The core operation of the transformer is self-attention, which allows tokens to exchange information based on their learned similarity. 
Given token representations collected in a matrix $\mathbf{H}$, self-attention computes
\begin{equation}
    \mathrm{Attention}(\mathbf{Q},\mathbf{K},\mathbf{V})
    =
    \mathrm{softmax}\!\left(\frac{\mathbf{Q}\mathbf{K}^\top}{\sqrt{d}}\right)\mathbf{V},
\end{equation}
where queries $\mathbf{Q}$, keys $\mathbf{K}$, and values $\mathbf{V}$ are learned linear projections of $\mathbf{H}$ and $d$ is the dimension of the query and key vectors.
In \emph{global} attention, each token can attend to all other tokens, which allows the model to represent broad spatial dependencies across the full atmospheric field. 
In \emph{windowed} attention, as used in Swin-style transformers \cite{liu2021swintransformerhierarchicalvision}, attention is computed only within local windows, which reduces computational cost and introduces a stronger locality bias. 
By shifting the window partition between layers, Swin attention still allows information to propagate across larger spatial scales while retaining much of the efficiency of local processing.

After several layers of attention and nonlinear transformations, the final token representations are mapped back to the original atmospheric grid through a decoding or \emph{de-tokenization} step,
\begin{equation}
    \hat{\mathbf{x}}_{t+\Delta t} = \mathrm{DeTokenize}(\mathbf{H}^{(L)}),
\end{equation}
where $\mathbf{H}^{(L)}$ denotes the token representations after the final transformer layer. 
This step reconstructs the predicted atmospheric field from the sequence of latent patch representations.

Transformer-based architectures have become increasingly common in data-driven weather and climate modelling, including 3D transformer models for global weather forecasting such as Pangu-Weather~\cite{bi2023pangu} and foundation models for Earth-system prediction such as Aurora~\cite{bodnar2025foundation}.

Relative to CNNs, transformers impose a weaker hard-coded locality bias and can in principle represent broader spatial dependencies more easily.
The distinction between global and windowed attention is therefore important: the former emphasizes unrestricted interactions, whereas the latter trades some of this flexibility for greater efficiency and a more local inductive bias.

\section{Experimental design and implementation details}
\label{app:models}
This section summarizes the emulator benchmark pipeline, including Isca data generation, model-specific preprocessing, details on machine learning architectures, training protocols, and evaluation.

\subsection{Implementation details of Isca configuration}
The two datasets are generated from the same idealized dry Isca configuration and differ only in whether the additional wave-heating perturbation is active. 
The no-SSW dataset omits this forcing, while the SSW-enabled dataset includes the prescribed wave-2 heating perturbation used to enhance planetary-wave activity and stratospheric variability. 
All other simulator settings, including horizontal resolution, vertical levels, time step, retained variables, spin-up removal, and train-validation-test split, are held fixed across the two datasets. 
The Isca configuration follows \cite{mudhar2024modeldependency}, with implementation details provided in the accompanying code repository. 
A compact summary of the simulator configuration and preprocessing choices is provided in Table~\ref{tab:isca_datasets}.

\subsection{Preprocessing of input representation}
\label{si:input_representations}
All input representations are constructed from the same raw Isca output fields. 
For each dataset, we retain temperature \(T\), zonal wind \(u\), and meridional wind \(v\) on the full pressure--latitude--longitude grid. 
The raw variables are stored with dimensions \((t,p,\phi,\lambda)\), and the preprocessing selects the same time window, variables, and train-validation-test split. 
Specifically, we discard the first 1000 output steps, retain 2500 consecutive states, and split these chronologically into training, validation, and test states. 
Normalization statistics are computed from the training split only, separately for each variable--pressure-level channel over time, latitude, and longitude, and then applied to the validation and test splits.
One-step prediction pairs \((\mathbf{x}_t,\mathbf{x}_{t+1})\) are formed within each split, so transitions across split boundaries are excluded.
The main difference between model families is therefore not the underlying physical data, but how the atmospheric state is represented for the neural network. 

\subsubsection{2D tensor representation}
\label{si:input_2d_tensor}
For the 2D convolutional and transformer models, pressure is flattened into the channel dimension rather than retained as an explicit spatial axis. 
Each Isca state is therefore stored as a latitude--longitude tensor with variables and pressure levels stacked as channels,
\(\mathbf{x}_t \in \mathbb{R}^{180 \times 64 \times 128}\), where \(180 = 3 \times 60\) corresponds to three prognostic variables on 60 pressure levels. 
This representation allows the models to operate directly on horizontal structure, while vertical interactions are represented through channel mixing.

The coordinate features are appended dynamically to the input at load time. 
They consist of four trigonometric latitude--longitude channels, \(\sin\phi\), \(\cos\phi\), \(\sin\lambda\), and \(\cos\lambda\). 
Thus, the runtime input has 184 channels on the latitude--longitude grid, while the target remains the 180 prognostic variable--level channels.

\subsubsection{3D tensor representation}
\label{si:input_3d_tensor}
For the 3D convolutional and transformer models, pressure is retained as an explicit spatial axis rather than flattened into channels. 
Each Isca state is therefore stored as a tensor with variables, pressure, latitude, and longitude dimensions,
\(\mathbf{x}_t \in \mathbb{R}^{3 \times 60 \times 64 \times 128}\). 
This representation allows the models to apply local operations directly across neighboring pressure levels as well as across latitude and longitude.

The coordinate features are appended dynamically to the input at load time. 
They consist of four trigonometric latitude--longitude channels,\(\sin\phi\), \(\cos\phi\), \(\sin\lambda\), and \(\cos\lambda\), together with two pressure-coordinate channels.
The pressure channels are constructed by min--max normalizing pressure \(p\) and log-pressure \(\log p\) across the retained pressure levels, and then broadcasting these values over latitude and longitude. 
Thus, the runtime input has 9 channels over the pressure--latitude--longitude volume, while the target remains the 3 prognostic variables.

\subsubsection{Grid MPNN representation}
\label{si:input_grid_gnn}
For the grid MPNN, each Isca state is represented as a graph on the native latitude--longitude grid. 
Each horizontal grid point is treated as a node, giving \(N=64\times128=8192\) nodes. 
At each node, the three prognostic variables across 60 pressure levels are stacked into a dynamic node-feature vector with \(3\times60=180\) features. 
The saved dynamic state tensor therefore has shape \(N_t \times N \times F_{\mathrm{dyn}} = N_t \times 8192 \times 180\), where \(N_t\) denotes the number of saved time steps, the second axis indexes graph nodes, and the last axis indexes dynamic variable--pressure-level features.

The graph connectivity is fixed across time.
Nodes are connected using a fixed 8-neighbor connectivity pattern on the latitude--longitude grid, with periodic wrapping in longitude and spherical wrapping across the poles. 
For neighbors that cross a pole, the implementation wraps the connection across the spherical grid by shifting the longitude by 180$^\circ$, so that the neighbor is placed consistently on the other side of the pole.
This gives a directed regular graph with \(8192\times8=65536\) edges.
Each edge is assigned four geometric features.
For an edge from sender node \(j\) to receiver node \(i\), these are the normalized Euclidean edge length and the three normalized components of the Cartesian displacement \(\mathbf{x}_j-\mathbf{x}_i\), expressed in a local coordinate basis at the receiver node.
This basis consists of the outward surface-normal direction and the local eastward and northward tangent directions.
All four quantities are normalized by the maximum edge length in the graph.

Static node-coordinate features are appended dynamically to the model input at load time. 
These consist of \(\sin\phi\), \(\cos\phi\), \(\sin\lambda\), and \(\cos\lambda\), increasing the runtime input from 180 to 184 node features, while the target remains the 180 prognostic variable--level features. 

\subsubsection{Mesh MPNN representation}
\label{si:input_mesh_gnn}
For the mesh MPNN, each Isca state is first represented on the native latitude--longitude grid with \(N_{\mathrm{grid}}=64\times128=8192\) grid nodes. 
At each grid node, the three prognostic variables across 60 pressure levels are stacked into a dynamic feature vector with \(3\times60=180\) features. 
The saved dynamic state tensor therefore has shape \(N_t \times N \times F_{\mathrm{dyn}} = N_t \times 8192 \times 180\), where \(N_t\) denotes the number of saved time steps, the second axis indexes graph nodes, and the last axis indexes dynamic variable--pressure-level features.

The mesh MPNN then uses a static spherical mesh as an intermediate representation. 
The model first maps information from the latitude--longitude grid to nearby mesh nodes, performs message passing on the mesh, and finally maps the mesh representation back to the original grid. 
The mesh is constructed by starting from an icosahedron and refining it four times, giving 2562 nodes at the finest level. 
The mesh-to-mesh graph uses connections from all refinement levels, so the processor can exchange information across multiple spatial scales rather than only between neighboring nodes on the finest mesh.

The static graph contains three edge sets. 
Grid-to-mesh edges connect each grid node to nearby finest-mesh nodes using a radius query in unit-sphere Cartesian space. 
For this configuration, the radius is set to \(0.8\) times the maximum edge length of the finest mesh. 
Mesh-to-mesh edges connect nodes within the multi-resolution mesh. 
Mesh-to-grid edges connect each grid node to the three vertices of the finest-mesh triangle that contains it. 
In this setup, the graph contains 22960 grid-to-mesh edges, 20460 mesh-to-mesh edges, and 24576 mesh-to-grid edges.

Static node-coordinate features are included for both grid and mesh nodes. 
They consist of Cartesian coordinates on the unit sphere, \(x\), \(y\), and \(z\), together with trigonometric latitude--longitude features, \(\sin\phi\), \(\cos\phi\), \(\sin\lambda\), and \(\cos\lambda\). 
Each edge is assigned four geometric features.
For an edge from sender node \(j\) to receiver node \(i\), these are the normalized Euclidean edge length and the three normalized components of the Cartesian displacement \(\mathbf{x}_j-\mathbf{x}_i\), expressed in a local coordinate basis at the receiver node.
This basis consists of the outward surface-normal direction and the local eastward and northward tangent directions.
All four quantities are normalized by the maximum edge length within the corresponding edge set.

\subsection{Selected model configurations}
\label{si:selected_model_configurations}
For each architecture and dataset, the final model used in the reported test-set results is the run with the lowest validation MAE from the corresponding hyperparameter sweep. 
The sweep spaces are summarized in Tables~\ref{tab:cnn_sweep_space}--\ref{tab:transformer_sweep_space}. 
The subsections below summarize the selected configurations for each architecture family, while the full run logs, validation histories, and complete module printouts are available at the public Weights \& Biases projects linked in the accompanying code repository.

\subsubsection{Simple CNN2D}
For the no-SSW dataset, the selected Simple CNN2D used a hidden dimension of 1024, three hidden convolutional layers, kernel size \(7\times7\), SiLU activations, lon-lat padding, and no batch normalization. 
Lon-lat padding means that longitude is padded circularly to respect periodicity, while latitude is padded by replicating the boundary values.
The model first projects the 184 input channels to 1024 hidden channels using a \(1\times1\) convolution, applies three \(7\times7\) convolutions over the latitude--longitude grid, and then maps back to the 180 prognostic target channels using a final \(1\times1\) convolution. 
It was trained with MSE loss, the Adam optimizer, learning rate \(10^{-4}\), weight decay \(10^{-4}\), batch size 5, and 50 epochs. 
The selected model had \(154.5\)M trainable parameters.

For the SSW-enabled dataset, the selected Simple CNN2D used the same hidden dimension, activation, padding, optimizer, loss, learning rate, weight decay, batch size, and number of epochs, but used six hidden convolutional layers with kernel size \(5\times5\). 
The selected SSW-enabled model had \(157.7\)M trainable parameters.

\subsubsection{CNN2D U-Net}
For the no-SSW dataset, the selected CNN2D U-Net used a base hidden dimension of 256, max-pooling downsampling, SiLU activations, lon-lat padding, no batch normalization, and a bottleneck dilation of 3. 
The model uses an encoder--decoder structure with skip connections, with channel widths increasing from 256 to 512 and 1024 in the downsampling path. 
Both CNN2D U-Net models use \(5\times5\) convolutions at the highest-resolution encoder level, \(3\times3\) convolutions at the lower-resolution encoder/decoder levels, and dilated \(3\times3\) convolutions in the bottleneck.
It was trained with L1 loss, the AdamW optimizer, learning rate \(10^{-3}\), weight decay \(10^{-4}\), batch size 5, and 50 epochs.
The selected model had approximately \(36\)M trainable parameters.

For the SSW-enabled dataset, the selected CNN2D U-Net used the same activation, padding, downsampling method, batch-normalization setting, loss, weight decay, batch size, and number of epochs, but increased the base hidden dimension to 512 and used a bottleneck dilation of 4. 
This gives channel widths of 512, 1024, and 2048 across the encoder. 
It was trained with the Adam optimizer and learning rate \(10^{-4}\).
The selected model had approximately \(147\)M trainable parameters.

\subsubsection{Simple CNN3D}
For the no-SSW dataset, the selected Simple CNN3D used a hidden dimension of 64, four hidden convolutional layers, kernel size \(7\times7\times7\), SiLU activations, lon-lat padding, and no batch normalization. 
The model first projects the 9 input channels to 64 hidden channels using a \(1\times1\times1\) convolution, applies four 3D convolutional layers over pressure, latitude, and longitude, and then maps back to the 3 prognostic target channels using a final \(1\times1\times1\) convolution. 
For 3D convolutions, lon-lat padding uses circular padding in longitude and replicated padding in latitude and pressure. 
The model was trained with MSE loss, the Adam optimizer, learning rate \(10^{-4}\), weight decay \(10^{-4}\), batch size 5, and 50 epochs. 
The selected model had \(5.6\)M trainable parameters.

For the SSW-enabled dataset, the selected Simple CNN3D used the same hidden dimension, number of layers, kernel size, padding, batch-normalization setting, learning rate, weight decay, batch size, number of epochs, and parameter count. 
The main differences were that it used GELU activations, L1 loss, and the AdamW optimizer.

\subsubsection{CNN3D U-Net}
For the no-SSW dataset, the selected CNN3D U-Net used a base hidden dimension of 40, strided-convolution downsampling, GELU activations, lon-lat padding, no batch normalization, and a bottleneck dilation of 3. 
The model uses an encoder--decoder structure with skip connections, with channel widths increasing from 40 to 80 and 160 in the downsampling path. 
Both CNN3D U-Net models use \(5\times5\times5\) convolutions at the highest-resolution encoder level, \(3\times3\times3\) convolutions at the lower-resolution encoder/decoder levels, and dilated \(3\times3\times3\) convolutions in the bottleneck.
It was trained with L1 loss, the Adam optimizer, learning rate \(10^{-3}\), weight decay \(10^{-4}\), batch size 5, and 50 epochs. 
The selected model had approximately \(3\)M trainable parameters.

For the SSW-enabled dataset, the selected CNN3D U-Net used the same activation, padding, batch-normalization setting, optimizer, weight decay, batch size, and number of epochs, but used a base hidden dimension of 50, mean-pooling downsampling, and a bottleneck dilation of 2. 
This gives channel widths of 50, 100, and 200 across the encoder. 
It was trained with SmoothL1 loss and learning rate \(5\times10^{-4}\). 
The selected model had approximately \(5\)M trainable parameters.

\subsubsection{Transformer 2D Global Attention}
The Transformer 2D Global Attention model uses the same two-dimensional latitude--longitude representation as the CNN2D models. 
The input consists of 180 prognostic channels, corresponding to three variables across 60 pressure levels, plus four trigonometric coordinate channels. 
The \(64\times128\) grid is partitioned into non-overlapping \(4\times2\) patches, which are embedded by a convolution with matching kernel size and stride. 
This gives a \(16\times64\) token grid, flattened into 1024 tokens. 
After adding learned positional embeddings, the tokens are processed by transformer blocks with pre-normalized eight-head global self-attention, MLP layers, and residual connections. 
The decoded tokens are mapped to \(180\times4\times2\) values, rearranged to the full spatial grid, and added as a residual update to the prognostic input channels.

For the no-SSW dataset, the selected model used 10 transformer blocks, hidden dimension 2048, MLP expansion ratio 6, GELU activations, and no dropout or attention dropout. 
It was trained with SmoothL1 loss, AdamW, learning rate \(3\times10^{-4}\), weight decay \(5\times10^{-4}\), batch size 5, and 50 epochs, giving \(679.5\)M trainable parameters.

For the SSW-enabled dataset, the selected model kept the same patch size, number of blocks, attention heads, activation, dropout settings, loss, optimizer, weight decay, batch size, and number of epochs, but used hidden dimension 1024, MLP expansion ratio 8, and learning rate \(5\times10^{-4}\). 
The selected model had \(213.9\)M trainable parameters.

\subsubsection{Transformer 2D Swin}
The Transformer 2D Swin model uses the same 184-channel two-dimensional input representation as the Transformer 2D Global Attention model. 
The \(64\times128\) grid is partitioned into non-overlapping \(2\times2\) patches, embedded by a convolution with matching kernel size and stride, producing a \(32\times64\) token grid with 2048 tokens. 
Learned positional embeddings are added before the tokens are processed by Swin transformer blocks adapted from the torchvision implementation (\url{https://github.com/pytorch/vision/blob/main/torchvision/models/swin_transformer.py}).
Instead of applying attention globally, each block applies multi-head self-attention within local windows, alternating between unshifted and shifted windows to exchange information across window boundaries. 
Attention uses learned relative-position biases, followed by MLP layers, pre-normalization, and residual connections. 
A linear decoder maps each token to \(180\times2\times2\) values, reconstructs the full grid, and adds the result as a residual update to the prognostic input state.

For the no-SSW dataset, the selected model used hidden dimension 1152, eight transformer blocks, eight attention heads, MLP expansion ratio 8, GELU activations, no dropout or attention dropout, and \(2\times4\)-token attention windows. 
It was trained with SmoothL1 loss, AdamW, learning rate \(10^{-3}\), weight decay \(5\times10^{-4}\), batch size 5, and 50 epochs. 
The selected model had \(216.5\)M trainable parameters.

For the SSW-enabled dataset, the selected model kept the same hidden dimension, patch size, attention heads, activation, dropout settings, optimizer, weight decay, batch size, and number of epochs, but used 12 transformer blocks, MLP expansion ratio 6, \(4\times4\)-token attention windows, L1 loss, and learning rate \(5\times10^{-4}\). 
The selected model had \(259.1\)M trainable parameters.

\subsubsection{Transformer 3D Global Attention}
The Transformer 3D Global Attention model is the direct three-dimensional analogue of the Transformer 2D Global Attention model. 
Instead of stacking pressure levels as channels, it operates on the full pressure--latitude--longitude grid. 
The input consists of three prognostic state channels and six coordinate channels: sine and cosine encodings of latitude and longitude, and normalized linear and logarithmic pressure coordinates. 
This gives nine input channels on a \(60\times64\times128\) grid.

As in the 2D global-attention model, a strided convolution embeds non-overlapping patches as tokens, learned positional embeddings are added, and ten transformer blocks apply pre-normalized eight-head global self-attention followed by MLP layers and residual connections. 
The only architectural difference is that patching, attention tokens, and decoding are defined over three-dimensional pressure--latitude--longitude patches. 
The decoder maps each processed token back to the three prognostic variables within its patch, reconstructs the full grid, and adds the result as a residual update to the input state.

For the no-SSW dataset, the selected model used \(4\times4\times4\) patches, giving a \(15\times16\times32\) token grid, or 7680 tokens. 
It used hidden dimension 768, MLP expansion ratio 6, GELU activations, and no dropout or attention dropout. 
It was trained with SmoothL1 loss, AdamW, learning rate \(10^{-3}\), weight decay \(5\times10^{-4}\), batch size 5, and 50 epochs. 
The selected model had \(101.0\)M trainable parameters.

For the SSW-enabled dataset, the selected model kept the same number of transformer blocks, attention heads, MLP expansion ratio, activation, dropout settings, optimizer, learning rate, weight decay, batch size, and number of epochs, but used hidden dimension 896, \(6\times4\times4\) patches, and L1 loss. 
This gives a \(10\times16\times32\) token grid, or 5120 tokens. 
The selected model had \(134.2\)M trainable parameters.

\subsubsection{Transformer 3D Swin}
The Transformer 3D Swin model is the three-dimensional analogue of the Transformer 2D Swin model. 
It operates directly on the pressure--latitude--longitude grid instead of stacking pressure levels as channels. 
The input consists of three prognostic state channels and six coordinate channels: sine and cosine encodings of latitude and longitude, and normalized linear and logarithmic pressure coordinates, giving nine input channels on a \(60\times64\times128\) grid.

As in the 3D global-attention transformer, a strided three-dimensional convolution embeds non-overlapping pressure--latitude--longitude patches as tokens, and a linear decoder reconstructs the full grid as a residual update to the input state. 
The difference is that attention is applied locally using three-dimensional shifted windows, adapted from the torchvision video Swin transformer implementation: \url{https://raw.githubusercontent.com/pytorch/vision/main/torchvision/models/video/swin_transformer.py}. 
Blocks alternate between unshifted and shifted windows, allowing information to propagate across window boundaries while keeping attention local in pressure, latitude, and longitude.

For the no-SSW dataset, the selected model used \(4\times4\times4\) patches, giving a \(15\times16\times32\) token grid. 
It used hidden dimension 768, eight transformer blocks, eight attention heads, MLP expansion ratio 6, GELU activations, no dropout or attention dropout, and \(3\times2\times4\)-token attention windows. 
It was trained with L1 loss, AdamW, learning rate \(10^{-3}\), weight decay \(5\times10^{-4}\), batch size 5, and 50 epochs. 
The selected model had \(82.1\)M trainable parameters.

For the SSW-enabled dataset, the selected model kept the same hidden dimension, patch size, number of blocks, attention heads, MLP expansion ratio, activation, dropout settings, learning rate, batch size, and number of epochs, but used \(3\times4\times2\)-token attention windows, SmoothL1 loss, and weight decay \(10^{-4}\). 
The selected model had \(82.1\)M trainable parameters.

\subsubsection{Grid MPNN}
The regular-grid MPNN 2D uses the grid-graph representation described in Section~\ref{si:input_grid_gnn}. 
It therefore operates on \(8192\) latitude--longitude nodes with 184 input features per node: 180 dynamic prognostic features and four static trigonometric coordinate features. 
The graph has fixed eight-neighbour directed connectivity with \(65{,}536\) edges, each carrying four geometric edge features.

The selected model is a gated graph convolutional network with six residual message-passing layers and hidden dimension 500. 
In each layer, sender-node, receiver-node, and edge embeddings are combined to update an edge gate, which controls the contribution of each neighbouring node before aggregation. 
A decoder predicts a 180-dimensional residual at each node, which is added to the current prognostic state.

For the no-SSW dataset, the selected configuration used SiLU activations, no dropout, a three-layer node encoder, a three-layer edge encoder, and a three-layer decoder. 
It was trained with Adam, learning rate \(5\times10^{-4}\), and had \(8.73\)M trainable parameters.

For the SSW-enabled dataset, the selected configuration used SiLU activations, dropout \(0.1\), a three-layer node encoder, a two-layer edge encoder, and a two-layer decoder. 
It was trained with AdamW, learning rate \(10^{-3}\), and had \(8.48\)M trainable parameters.

\subsubsection{Mesh MPNN}
The mesh-based MPNN 2D uses the grid--mesh representation described in Section~\ref{si:input_mesh_gnn}. 
It follows a grid-to-mesh, mesh-to-mesh, and mesh-to-grid architecture, where information is first transferred from the native latitude--longitude grid to the spherical mesh, processed on the multi-resolution mesh graph, and then mapped back to the original grid.

Separate MLP encoders map grid-node features, mesh-node features, and the three edge-feature sets into a common hidden dimension. 
The mesh processor then applies gated graph convolutional layers over the multi-resolution mesh graph. 
For each edge, sender-node, receiver-node, and edge embeddings are combined to update an edge gate, which weights the sender message before aggregation. 
Residual connections are used for both node and edge updates. 
After the mesh-to-grid stage, an MLP decoder predicts a 180-dimensional residual update at each grid point, which is added to the current prognostic state.

For the no-SSW dataset, the selected model used hidden dimension 640, two grid-to-mesh layers, nine mesh-to-mesh layers, and two mesh-to-grid layers. 
It used LeakyReLU activations, dropout \(0.1\), two-layer node encoders, three-layer edge encoders, and a three-layer decoder. 
The selected model had \(31.1\)M trainable parameters.

For the SSW-enabled dataset, the selected model used hidden dimension 512, three grid-to-mesh layers, nine mesh-to-mesh layers, and two mesh-to-grid layers. 
It used SiLU activations, no dropout, two-layer node encoders, three-layer edge encoders, and a three-layer decoder. 
The selected model had \(21.2\)M trainable parameters.

\subsection{Training budget and computational constraints}
\label{si:training_budget}
To make the architecture comparison as controlled as possible, all model families were trained under a shared one-step forecasting protocol and a fixed hyperparameter-sweep budget. 
For each architecture and dataset, we ran a Bayesian sweep with a run cap of 50 configurations and selected the model with the lowest validation MAE. 
The corresponding search spaces are reported in Tables~\ref{tab:cnn_sweep_space}--\ref{tab:transformer_sweep_space}.

The goal of this setup is not to optimize each architecture family to its absolute performance limit, but to compare representative implementations under comparable tuning effort. 
This is important because the model classes differ substantially in parameter count, memory footprint, and optimization behavior. 
Using the same sweep budget reduces the risk that observed differences are driven mainly by unequal architecture-specific tuning effort rather than by the inductive biases being tested.

All selected models were trained for 50 epochs with batch size 5, mixed-precision training, and a fixed random seed. 
Training was performed on a single GPU per run, primarily using an NVIDIA RTX 5080. 

\clearpage

\begin{figure}
    \centering
    \includegraphics[width=\linewidth]{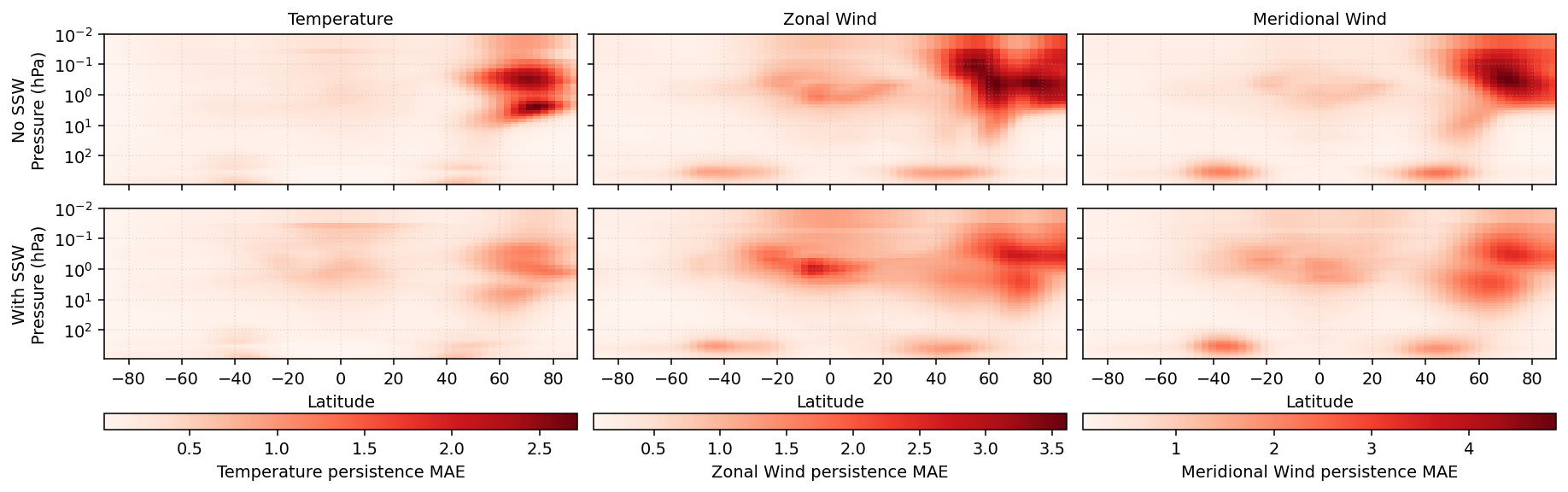}
    \caption{Pressure--latitude structure of persistence mean absolute error for temperature, zonal wind, and meridional wind in the no-SSW and SSW-enabled regimes in the hold-out test set. 
    Persistence errors are substantial in both regimes, showing that even over the 3-hour prediction horizon the grid-point fields undergo appreciable local evolution. 
    The largest errors are concentrated in the Northern Hemisphere stratosphere and are generally stronger in the no-SSW regime, especially for the wind components. 
    This likely reflects the stronger and more coherent polar vortex in the unforced configuration, where small displacements of sharp wind gradients can produce large grid-point errors. 
    Persistence provides a useful reference for short-term local variability, but should not be interpreted as a complete measure of dynamical difficulty.}
    \label{fig:si_persistance_mae_lat_plevels_allvars}
\end{figure}

\clearpage

\begin{figure}
    \centering
    \includegraphics[width=0.47\linewidth]{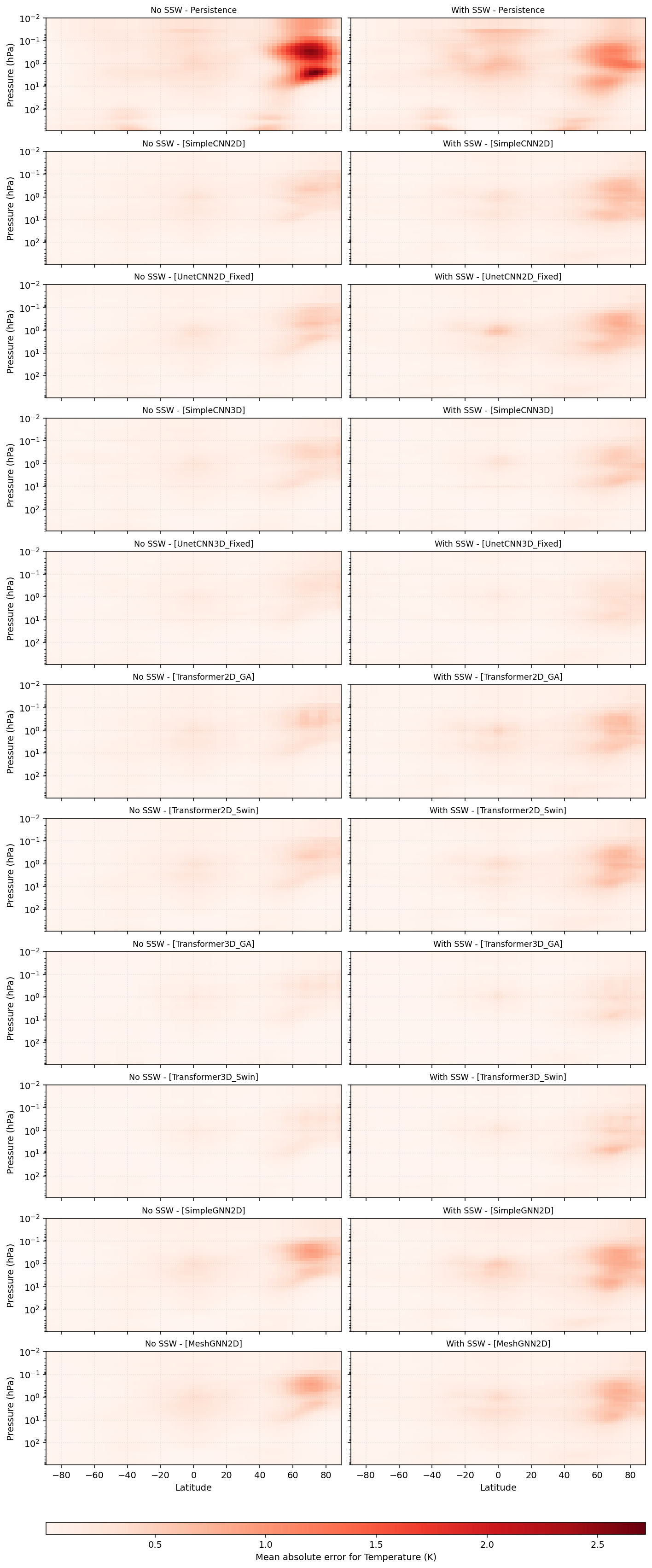}
    \caption{Latitude--pressure structure of temperature mean absolute error for all emulator architectures. This diagnostic complements the main-text by showing how prediction errors vary with latitude and pressure for the temperature component.}
    \label{fig:mae_plevel_lat_all_models_temp}
\end{figure}

\clearpage

\begin{figure}
    \centering
    \includegraphics[width=0.47\linewidth]{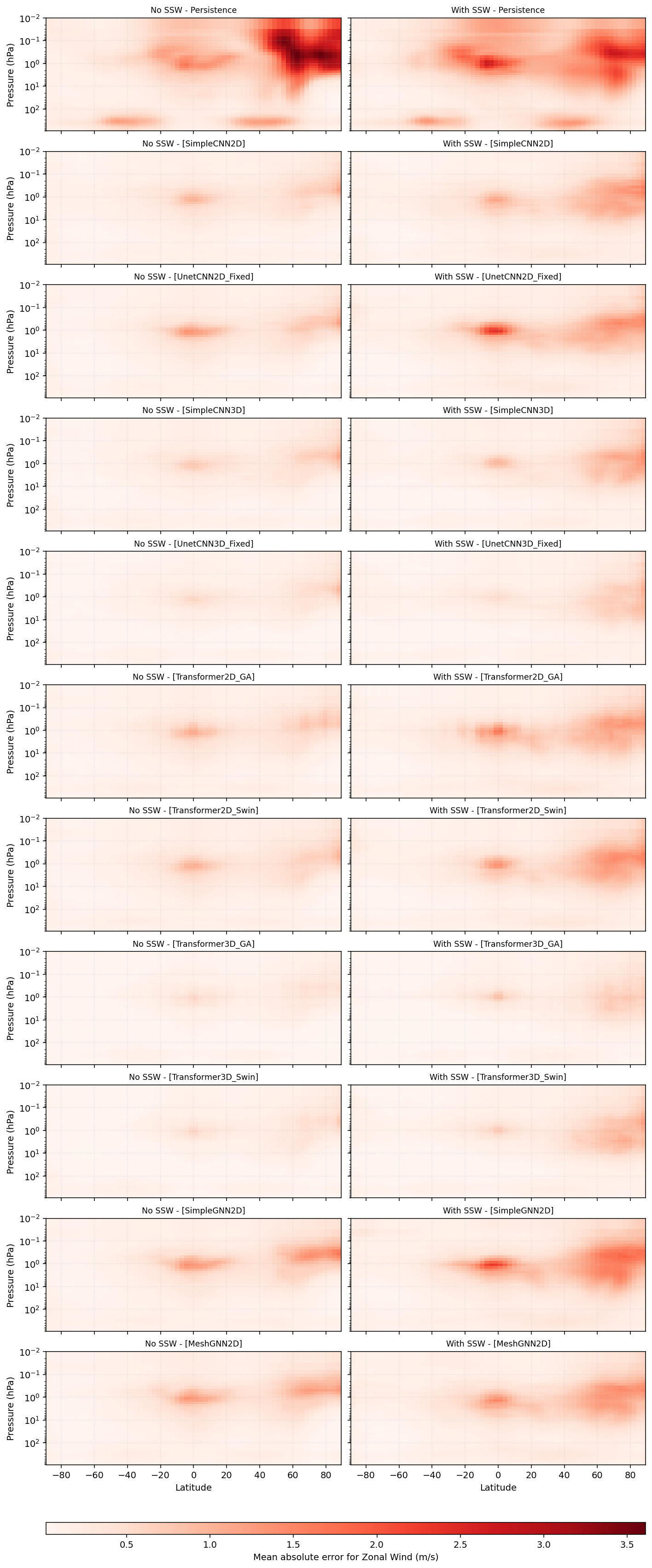}
    \caption{Latitude--pressure structure of zonal wind mean absolute error for all emulator architectures. This diagnostic complements the main-text by showing how prediction errors vary with latitude and pressure for the zonal wind component.}    
    \label{fig:mae_plevel_lat_all_models_vwind}
\end{figure}

\clearpage

\begin{figure}
    \centering
    \includegraphics[width=0.47\linewidth]{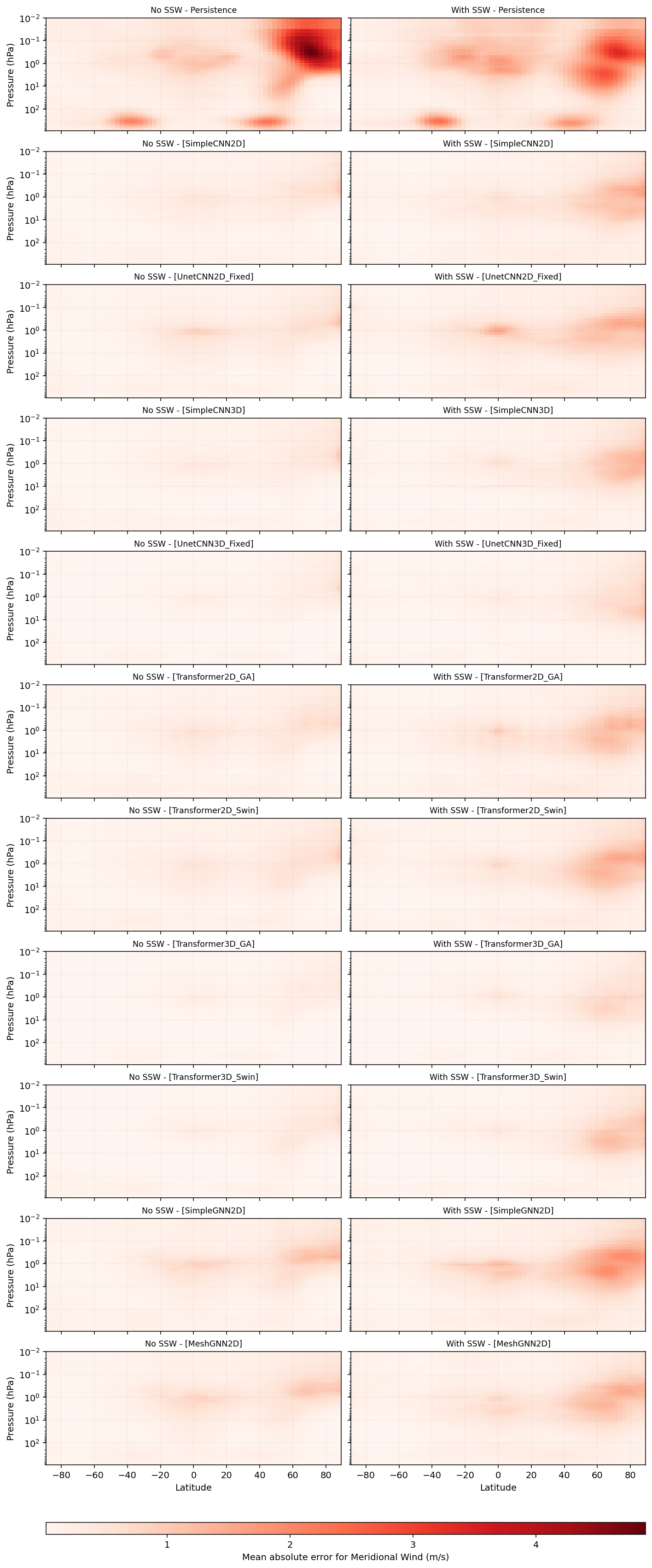}
    \caption{Latitude--pressure structure of meridional wind mean absolute error for all emulator architectures. This diagnostic complements the main-text by showing how prediction errors vary with latitude and pressure for the meridional wind component.}
    \label{fig:mae_plevel_lat_all_models_uwind}
\end{figure}

\clearpage

\begin{figure}
    \centering
    \includegraphics[width=0.95\linewidth]{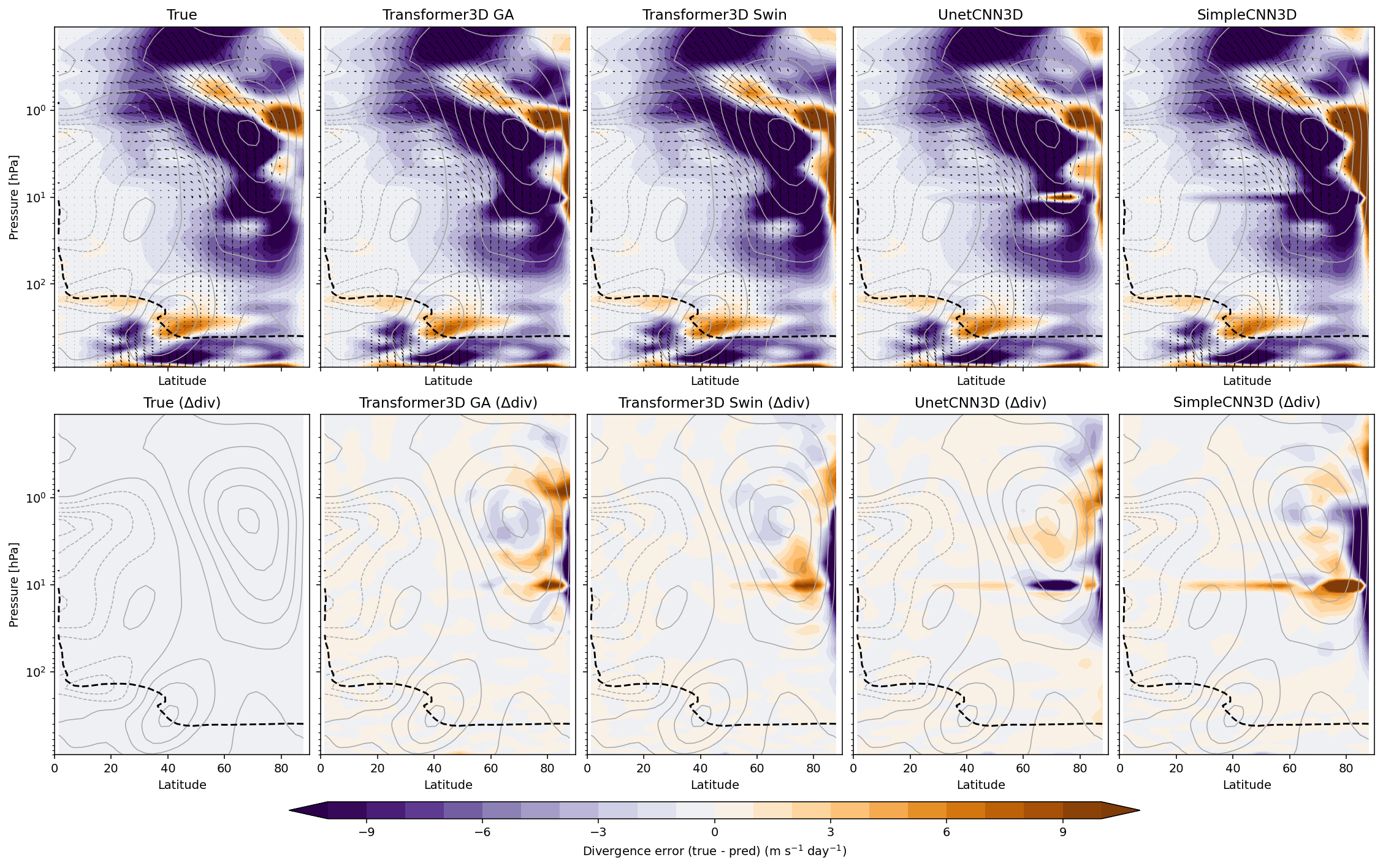}
    \caption{EP-flux diagnostics for 3D architectures averaged over the 10 days preceding the held-out SSW event described in section 3.3 in the main text.
    The upper row shows EP-flux vectors overlaid on EP-flux divergence for the true simulator trajectory and model predictions.
    The lower row shows divergence error, defined as true minus predicted divergence.
    The dashed contour marks the dynamical tropopause, defined here as the zonal- and time-mean contour of absolute potential vorticity.}   
    \label{fig:epflux3d}
\end{figure}

\clearpage

\begin{figure}[h]
    \centering
    \includegraphics[width=0.95\linewidth]{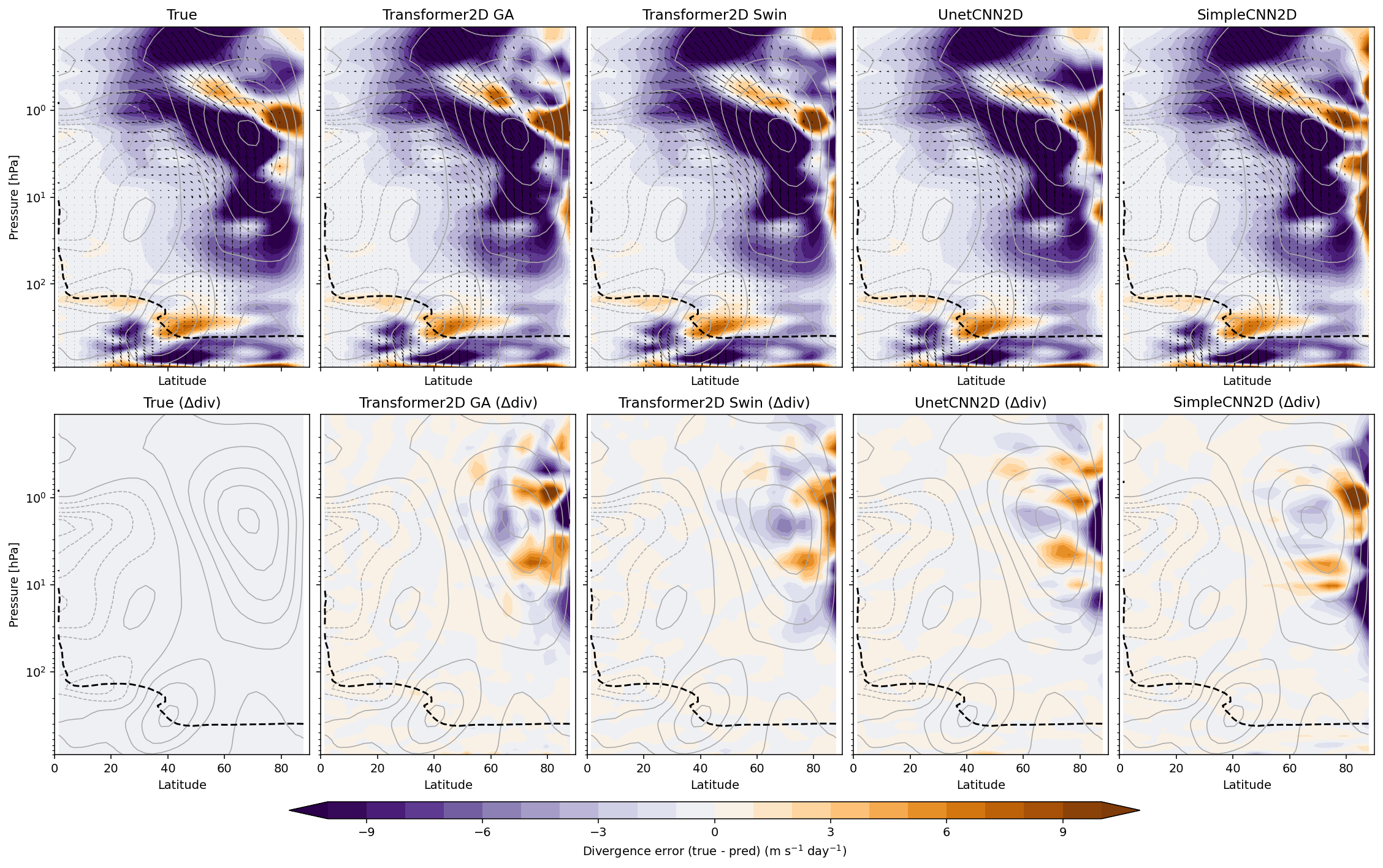}
    \caption{EP-flux diagnostics for 2D architectures averaged over the 10 days preceding the held-out SSW event described in section 3.3 in the main text.
    The upper row shows EP-flux vectors overlaid on EP-flux divergence for the true simulator trajectory and model predictions.
    The lower row shows divergence error, defined as true minus predicted divergence.
    The dashed contour marks the dynamical tropopause, defined here as the zonal- and time-mean contour of absolute potential vorticity.}
    \label{fig:epflux2d}
\end{figure}

\clearpage

\begin{figure}[h]
    \centering
    \includegraphics[width=0.75\linewidth]{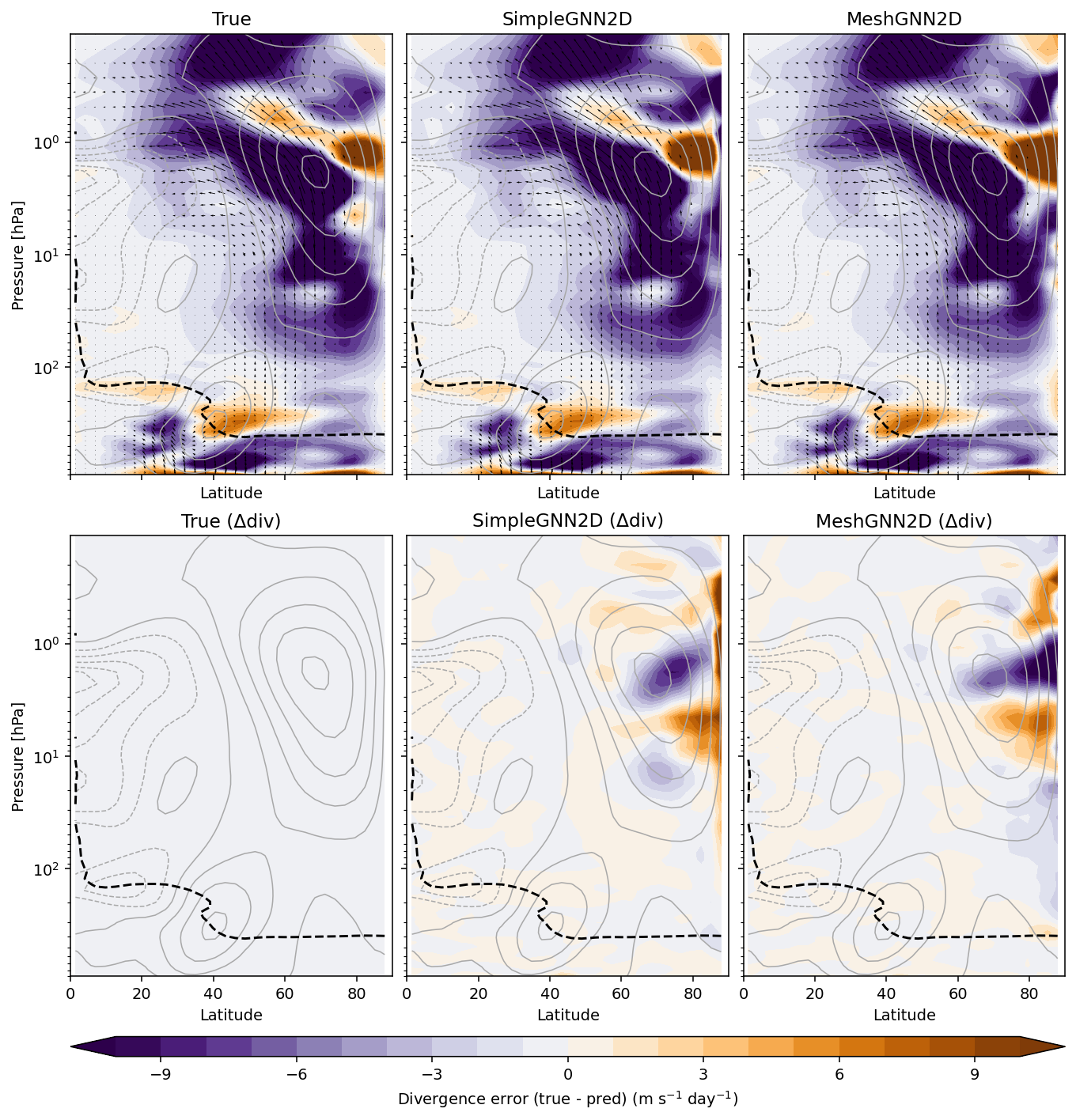}
    \caption{EP-flux diagnostics for graph-based architectures averaged over the 10 days preceding the held-out SSW event described in section 3.3 in the main text.
    The upper row shows EP-flux vectors overlaid on EP-flux divergence for the true simulator trajectory and model predictions.
    The lower row shows divergence error, defined as true minus predicted divergence.
    The dashed contour marks the dynamical tropopause, defined here as the zonal- and time-mean contour of absolute potential vorticity.}
    \label{fig:epfluxgnn}
\end{figure}

\clearpage

\begin{figure}[h]
    \centering
    \includegraphics[width=0.7\linewidth]{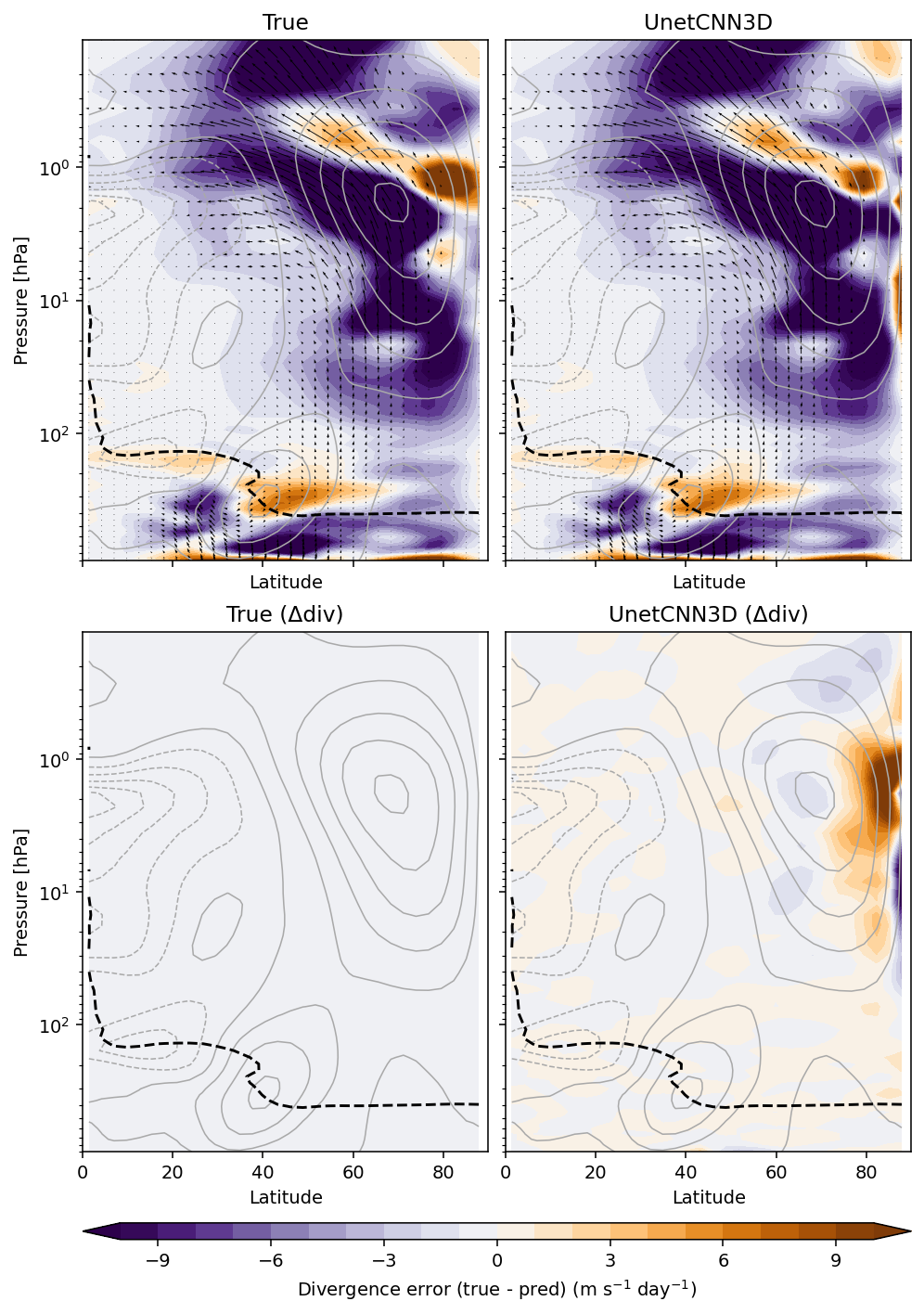}
    \caption{EP-flux diagnostics for the 3D U-Net after removing the explicit 10 hPa input level, which otherwise lies close to an existing 10.24 hPa level and causes the interesting artifact shown in the main text.
    The upper panels show EP-flux vectors overlaid on EP-flux divergence for the true simulator trajectory and model prediction, while the lower panels show the divergence error. 
    The narrow 10 hPa error band seen in Figure~\ref{fig:epflux3d} disappears, supporting its interpretation as a discretization-related artifact rather than a dynamical EP-flux error.}
    \label{fig:epflux_no_artifact}
\end{figure}

\clearpage

\begin{figure}[h]
    \centering
    \includegraphics[width=0.7\linewidth]{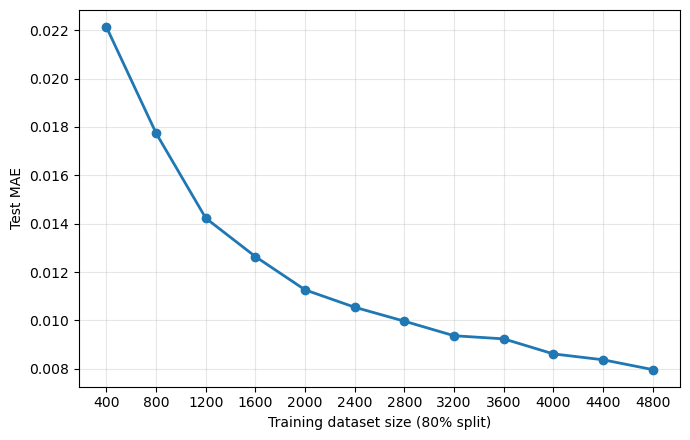}
    \caption{Data-scaling experiment for the selected training configuration. Test MAE decreases monotonically with training set size, with diminishing improvements at larger sample sizes. The training set size of 2000 was chosen to make it computationally feasible to experiment and hyperparameter-tune across many architectures.} 
\end{figure}

\clearpage


\begin{table}
\centering
\caption{Summary of the \textsc{Isca} configuration and dataset construction used for emulator training. The two datasets share the same base simulator configuration and differ only in whether the additional wave-heating perturbation is active.}
\label{tab:isca_datasets}
\resizebox{\columnwidth}{!}{%
\begin{tabular}{p{3.3cm} p{3.6cm} p{11.0cm}}
\cline{2-3}
 & \textbf{Attribute} & \textbf{Configuration} \\
\hline
\textbf{\textsc{Isca} simulation} & Base simulator & \textsc{Isca} dry spectral dynamical core with Held--Suarez troposphere and Polvani--Kushner stratosphere~\cite{held1994dynamicalcores,polvani2002stratosphere} \\
 & Horizontal resolution & T42 ($64 \times 128$ latitude--longitude equivalent) \\
 & Vertical levels & 60 pressure levels spanning 0.02--900 hPa; 38 levels above 100 hPa \\
 & Simulation length & $\sim$6 years \\
 & No-SSW dataset & Additional wave forcing absent \\
 & SSW-enabled dataset & Additional wave forcing present~\cite{mudhar2024modeldependency} \\
\hline
\textbf{Data processing} & Variables retained & $T$, $u$, $v$ \\
 & Sampling interval & 3 hours \\
 & Spin-up removal & First 1000 3-hourly time steps removed \\
 & Prediction pairs & $(\mathbf{x}_t,\mathbf{x}_{t+3h})$ \\
 & Data split & Chronological $2500$ training, $249$ validation, $249$ test samples\\
 & Normalization & Per-variable and per-pressure-level normalization using training statistics \\
\hline
\end{tabular}%
}
\end{table}

\clearpage

\begin{table}
\centering
\caption{Model families included in the benchmark and the main inductive-bias dimension each is intended to test.}
\label{tab:model_families}
\resizebox{\textwidth}{!}{%
\begin{tabular}{p{2.8cm} p{7.0cm} p{2.9cm} p{4.5cm} p{2.6cm}}
\hline
\textbf{Model family} & \textbf{Input} & \textbf{Model} & \textbf{Inductive bias} & \textbf{Explicit vertical coupling} \\
\hline
Baseline & Full atmospheric state & Persistence & Identity forecast & No \\
\hline
2D convolution & Latitude--longitude tensor with pressure levels and variables as channels & SimpleCNN2D & Local horizontal convolutions & No \\
\cmidrule(lr){3-5}
 &  & U-Net2D & Local multi-scale horizontal convolutions & No \\
\hline
3D convolution & Latitude--longitude--pressure tensor with variables as channels & SimpleCNN3D & Local 3D convolutions & Yes \\
\cmidrule(lr){3-5}
 &  & U-Net3D & Local multi-scale 3D convolutions & Yes \\
\hline
2D transformer & Latitude--longitude tensor with pressure levels and variables as channels. 2D patches used for tokenization. & Global ViT-2D & Global self-attention & No \\
\cmidrule(lr){3-5}
 &  & Swin ViT-2D & Windowed self-attention & No \\
\hline
3D transformer & Latitude--longitude--pressure tensor with variables as channels. 3D patches used for tokenization. & Global ViT-3D & Global self-attention & Yes \\
\cmidrule(lr){3-5}
 &  & Swin ViT-3D & Windowed self-attention & Yes \\
\hline
Grid GNN & Regular-grid graph with pressure levels and variables as node features and relative position and distance as edge features & Encoder--MPNN--decoder & Message passing on regular-grid graph & No \\
\hline
Mesh GNN & Grid-to-mesh-to-grid representation with pressure levels and variables stacked in node features and relative position and distance as edge features & GraphCast-style mesh model & Message passing on spherical mesh & No \\
\hline
\end{tabular}%
}
\end{table}

\clearpage

\begin{table}
\centering
\caption{Hyperparameter search spaces for the convolutional models. 
The same search spaces were used for both the SSW-enabled and no-SSW datasets. 
All sweeps used Bayesian optimization with a run cap of 50 and selected the configuration with the lowest validation MAE.}
\label{tab:cnn_sweep_space}
\scriptsize
\renewcommand{\arraystretch}{1.18}
\setlength{\tabcolsep}{3pt}
\begin{tabularx}{\textwidth}{p{2cm} p{1.2cm} p{1.2cm} p{2.1cm} p{1.1cm} p{1.7cm} p{1.5cm} p{1.7cm} p{1.7cm}}
\toprule
\textbf{Model} 
& \textbf{Hidden dim.} 
& \textbf{Num. layers} 
& \textbf{Convolution} 
& \textbf{Batch norm} 
& \textbf{Activation} 
& \textbf{Loss} 
& \textbf{Optimizer} 
& \textbf{Learning rate} \\
\midrule

SimpleCNN2D
& \makecell[l]{128\\256\\512\\1024}
& \makecell[l]{2\\3\\4\\5\\6}
& \makecell[l]{Kernel size:\\3\\5\\7}
& \makecell[l]{true\\false}
& \makecell[l]{GELU\\ReLU\\SiLU}
& \makecell[l]{L1\\SmoothL1\\MSE}
& \makecell[l]{AdamW\\Adam\\Yogi}
& \makecell[l]{$10^{-3}$\\$5{\times}10^{-4}$\\$10^{-4}$} \\

\midrule

U-Net2D
& \makecell[l]{Base:\\64\\128\\256\\512}
& --
& \makecell[l]{Bottleneck\\dilation:\\1, 2, 3, 4\\[0.25em]Downsampling:\\strided conv\\max pool\\mean pool}
& \makecell[l]{true\\false}
& \makecell[l]{GELU\\ReLU\\SiLU}
& \makecell[l]{L1\\SmoothL1\\MSE}
& \makecell[l]{AdamW\\Adam\\Yogi}
& \makecell[l]{$10^{-3}$\\$5{\times}10^{-4}$\\$10^{-4}$} \\

\midrule

SimpleCNN3D
& \makecell[l]{16\\32\\64}
& \makecell[l]{2\\3\\4}
& \makecell[l]{Kernel size:\\3\\5\\7}
& \makecell[l]{true\\false}
& \makecell[l]{GELU\\SiLU\\ReLU}
& \makecell[l]{L1\\SmoothL1\\MSE}
& \makecell[l]{AdamW\\Adam\\Yogi}
& \makecell[l]{$10^{-3}$\\$5{\times}10^{-4}$\\$10^{-4}$} \\

\midrule

U-Net3D
& \makecell[l]{Base:\\16\\32\\40\\50}
& --
& \makecell[l]{Bottleneck\\dilation:\\1, 2, 3\\[0.25em]Downsampling:\\mean pool\\max pool\\strided conv}
& \makecell[l]{true\\false}
& \makecell[l]{SiLU\\GELU\\LeakyReLU}
& \makecell[l]{L1\\SmoothL1\\MSE}
& \makecell[l]{AdamW\\Adam\\Yogi}
& \makecell[l]{$10^{-3}$\\$5{\times}10^{-4}$\\$10^{-4}$} \\
\bottomrule
\end{tabularx}
\end{table}

\clearpage

\begin{table}
\centering
\caption{Hyperparameter search spaces for the graph-based models. 
The same search spaces were used for both the SSW-enabled and no-SSW datasets. 
All sweeps used Bayesian optimization with a run cap of 50 and selected the configuration with the lowest validation MAE.}
\label{tab:gnn_sweep_space}
\scriptsize
\renewcommand{\arraystretch}{1.18}
\setlength{\tabcolsep}{3pt}
\begin{tabularx}{\textwidth}{p{1.0cm} p{1.2cm} p{1.0cm} p{1.0cm} p{1.0cm} p{1.7cm} p{1.1cm} p{1.3cm} p{1.4cm} p{1.5cm} p{1.5cm}}
\toprule
\textbf{Model} 
& \textbf{Hidden dim.} 
& \textbf{Node encoder} 
& \textbf{Edge encoder} 
& \textbf{De-coder} 
& \textbf{Message passing} 
& \textbf{Drop-out} 
& \textbf{Activa-tion} 
& \textbf{Loss} 
& \textbf{Optimizer}
& \textbf{Learning rate} \\
\midrule

Grid GNN
& \makecell[l]{256\\300\\350\\400\\450\\500}
& \makecell[l]{2\\3}
& \makecell[l]{2\\3}
& \makecell[l]{2\\3}
& \makecell[l]{GNN layers:\\4\\6\\8}
& \makecell[l]{0\\0.1}
& \makecell[l]{SiLU\\GELU\\Leaky-\\ReLU}
& \makecell[l]{L1\\SmoothL1\\MSE}
& \makecell[l]{AdamW\\Adam}
& \makecell[l]{$2.5{\times}10^{-3}$\\$10^{-3}$\\$5{\times}10^{-4}$\\$2{\times}10^{-4}$} \\

\midrule

Mesh GNN
& \makecell[l]{256\\512\\640\\768}
& \makecell[l]{2\\3}
& \makecell[l]{2\\3}
& \makecell[l]{2\\3}
& \makecell[l]{Grid--mesh:\\2\\3\\[0.25em]Mesh--mesh:\\5\\7\\8\\9\\10\\[0.25em]Mesh--grid:\\2\\3}
& \makecell[l]{0\\0.1}
& \makecell[l]{GELU\\ReLU\\SiLU\\Leaky-\\ReLU}
& \makecell[l]{L1\\SmoothL1\\MSE}
& \makecell[l]{AdamW\\Adam}
& \makecell[l]{$5{\times}10^{-3}$\\$2.5{\times}10^{-3}$\\$10^{-3}$\\$5{\times}10^{-4}$} \\

\bottomrule
\end{tabularx}
\end{table}

\clearpage

\begin{table}
\centering
\caption{Hyperparameter search spaces for the transformer models. 
The same search spaces were used for both the SSW-enabled and no-SSW datasets. 
All sweeps used Bayesian optimization with a run cap of 50 and selected the configuration with the lowest validation MAE.}
\label{tab:transformer_sweep_space}
\scriptsize
\renewcommand{\arraystretch}{1.18}
\setlength{\tabcolsep}{3pt}
\begin{tabularx}{\textwidth}{p{1.2cm} p{1.3cm} p{1.1cm} p{1.1cm} p{1.0cm} p{1.2cm} p{1.4cm} p{1.5cm} p{1.5cm} p{1.3cm} p{1.4cm}}
\toprule
\textbf{Model} 
& \textbf{Hidden dim.} 
& \textbf{Num. layers} 
& \textbf{Num. heads} 
& \textbf{MLP ratio} 
& \textbf{Patch size} 
& \textbf{Window size} 
& \textbf{Loss} 
& \textbf{Optimizer}
& \textbf{Learning rate}
& \textbf{Weight decay} \\
\midrule

Global ViT-2D
& \makecell[l]{512\\1024\\1280\\1536\\2048}
& \makecell[l]{4\\6\\8\\10\\12}
& \makecell[l]{4\\8}
& \makecell[l]{3\\4\\6\\8}
& \makecell[l]{Height:\\4\\8\\[0.25em]Width:\\2\\4\\8}
& --
& \makecell[l]{SmoothL1\\L1}
& AdamW
& \makecell[l]{$10^{-4}$\\$3{\times}10^{-4}$\\$5{\times}10^{-4}$\\$10^{-3}$}
& \makecell[l]{$10^{-4}$\\$5{\times}10^{-4}$} \\

\midrule

Swin ViT-2D
& \makecell[l]{768\\896\\1024\\1152\\1280\\1536\\1792\\2048}
& \makecell[l]{8\\10\\12}
& \makecell[l]{4\\8}
& \makecell[l]{6\\8}
& \makecell[l]{Height:\\2\\4\\[0.25em]Width:\\2\\4}
& \makecell[l]{Height:\\2\\4\\[0.25em]Width:\\2\\4}
& \makecell[l]{SmoothL1\\L1}
& AdamW
& \makecell[l]{$3{\times}10^{-4}$\\$5{\times}10^{-4}$\\$10^{-3}$}
& \makecell[l]{$10^{-4}$\\$5{\times}10^{-4}$} \\

\midrule

Global ViT-3D
& \makecell[l]{640\\768\\896\\1024\\1280}
& \makecell[l]{8\\10\\12}
& --
& \makecell[l]{4\\6}
& \makecell[l]{Depth:\\4\\5\\6\\[0.25em]Height:\\4\\8\\[0.25em]Width:\\4\\8}
& --
& \makecell[l]{SmoothL1\\L1}
& AdamW
& \makecell[l]{$10^{-4}$\\$3{\times}10^{-4}$\\$5{\times}10^{-4}$\\$10^{-3}$}
& \makecell[l]{$10^{-4}$\\$5{\times}10^{-4}$} \\

\midrule

Swin ViT-3D
& \makecell[l]{768\\896\\1024}
& \makecell[l]{8\\10\\12}
& --
& \makecell[l]{6\\8}
& \makecell[l]{Depth:\\4\\5\\[0.25em]Height:\\4\\8\\[0.25em]Width:\\4\\8}
& \makecell[l]{Height:\\2\\4\\[0.25em]Width:\\2\\4}
& \makecell[l]{SmoothL1\\L1}
& AdamW
& \makecell[l]{$3{\times}10^{-4}$\\$5{\times}10^{-4}$\\$10^{-3}$}
& \makecell[l]{$10^{-4}$\\$5{\times}10^{-4}$} \\

\bottomrule
\end{tabularx}
\end{table}

\end{document}